\documentclass[review]{article}

\PassOptionsToPackage{numbers, compress}{natbib}
 \usepackage[main, final]{neurips_2026}
\usepackage[utf8]{inputenc} % allow utf-8 input
\usepackage[T1]{fontenc}    % use 8-bit T1 fonts
\usepackage{hyperref}       % hyperlinks
\hypersetup{
    colorlinks=true,     % Use colored text for links instead of boxes
    linkcolor=blue,      % Color for internal links (e.g., equations, figures, sections)
    citecolor=blue,      % Color for bibliographic citations
    urlcolor=blue,       % Color for URLs/web links
    filecolor=blue       % Color for local file links
}
\usepackage{url}            % simple URL typesetting
\usepackage{booktabs}       % professional-quality tables
\usepackage{amsfonts}       % blackboard math symbols
\usepackage{nicefrac}       % compact symbols for 1/2, etc.
\usepackage{microtype}      % microtypography
\usepackage[table,dvipsnames]{xcolor} % Load once, with all needed options
\usepackage{tcolorbox}
\usepackage{enumitem}
\usepackage{subfig}
\usepackage{array}
\usepackage{booktabs}
\usepackage{pdfpages}
\usepackage{multirow}
\usepackage{supertabular, longtable}
\usepackage{amsmath}
\usepackage{soul}
\usepackage{wrapfig}

\definecolor{lightblue}{HTML}{d9ecff}
\definecolor{midblue}{HTML}{b3d9ff}
\definecolor{deepblue}{HTML}{7da4ef}
\definecolor{green}{HTML}{b3e5a1}
\definecolor{qblue}{RGB}{230,245,255}

\tcbuselibrary{skins,breakable}

\newcommand{\std}[1]{{\scriptsize\,(#1)}}

\title{Rep2Text: Decoding Full Text from a Single LLM Token Representation}

\author{
  Haiyan Zhao$^{1}$ \quad
  Zirui He$^{1}$ \quad
  Yiming Tang$^{2}$ \quad
  Fan Yang$^{3}$ \quad
  Ali Payani$^{4}$ \\
  {\bf Dianbo Liu}$^{2}$ \quad
  {\bf Mengnan Du}$^{5}$\textsuperscript{†}\\
  $^{1}$New Jersey Institute of Technology \quad
  $^{2}$National University of Singapore \quad
  $^{3}$Wake Forest University \\
  $^{4}$Cisco Research \quad
  $^{5}$The Chinese University of Hong Kong, Shenzhen \\
  \texttt{\{hz54,zh296\}@njit.edu} \quad
  \texttt{\{yiming,dianbo\}@nus.edu.sg} \\
  \texttt{yangfan@wfu.edu} \quad
  \texttt{apayani@cisco.com} \quad
  \texttt{mengnandu@cuhk.edu.cn}\\
  \textsuperscript{†}Corresponding author
}

\begin{document}

\maketitle
\vspace{-1em}

\begin{abstract}
Large language models (LLMs) have achieved remarkable progress across diverse tasks, yet their internal mechanisms remain largely opaque. In this work, we investigate a fundamental question: to what extent can the original input text be recovered from a single last-token representation in an LLM? To this end, we propose Rep2Text, a novel framework for decoding text from last-token representations. Rep2Text employs a trainable adapter that maps a target model's last-token representation into the token embedding space of a decoding language model, which then autoregressively reconstructs the input text. On Wikipedia-derived 16-token sequences, Rep2Text recovers roughly half of the tokens from a single last-token representation across multiple target and decoding model combinations, while preserving strong semantic coherence. Further analysis reveals a clear information bottleneck effect: as sequence length increases, token-level recovery declines, while semantic information remains relatively well preserved. We also find that decoder family matters more than decoder scale. Finally, our framework shows partial generalization to out-of-distribution clinical data. Our code is available at \url{https://github.com/hy-zhao23/Rep2Text}.
\end{abstract}

\section{Introduction}\label{sec:intro}
Large language models (LLMs) have achieved significant progress across a wide array of tasks. Despite their impressive performance, these models are often regarded as ``black boxes'', limiting our understanding of their internal mechanisms. Consequently, a growing body of research has sought to decode the information encoded in LLMs. These approaches vary widely, ranging from training linear probes~\citep{zou2023representation,gurneelanguage} or sparse autoencoders (SAEs)~\citep{shu2025survey} to interpret specific features, to mapping internal representations directly to the vocabulary space through methods like Logit Lens~\citep{nostalgebraist2020logitlens} and Tuned Lens~\citep{belrose2023eliciting}. In this work, we focus on a distinct but related challenge: \emph{representation decoding}, which aims to reconstruct the full original text from the internal representations of language models.

Building on this perspective, we ask: {\ul{To what extent can we recover the information contained in the last-token representation of an input sequence?}} Our overall goal is to explore single-token, activation-based input inversion. This is particularly challenging because the last-token representation is optimized for next-token prediction and can therefore be viewed as an information bottleneck. Through quantitatively comparing the inverted text with the original input, we aim to provide insight into what knowledge is preserved and encoded in the last-token representation of LLMs. 
%This question is critical in our community since we usually take the last-token representation as the representation of a sequence in decoder-only transformer models. However, 

\begin{figure*}[t]
    \centering
    \includegraphics[width=\linewidth]{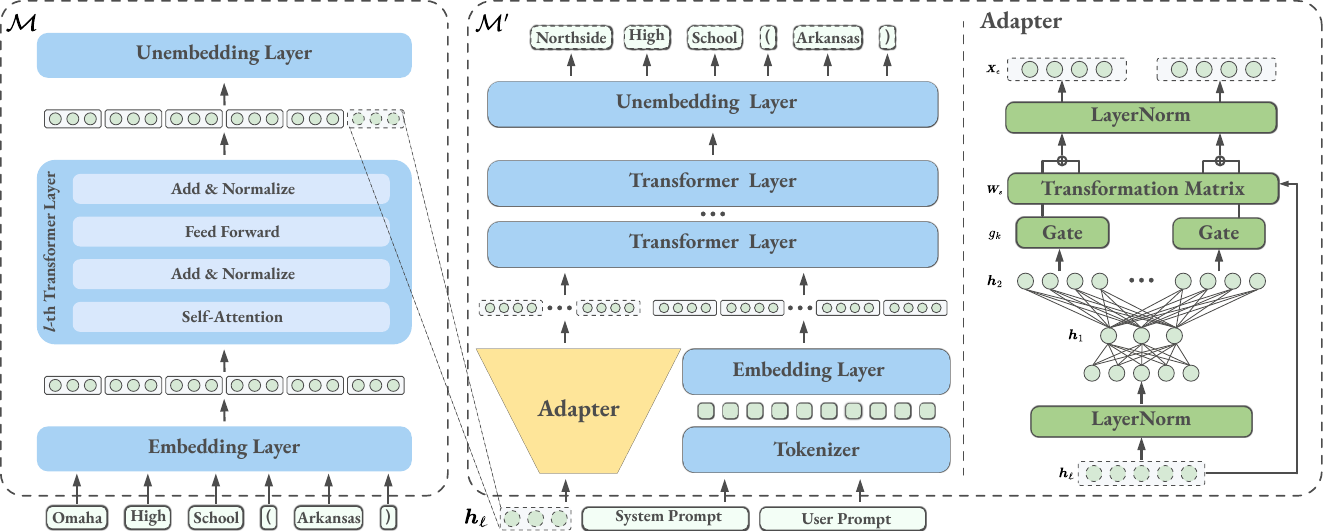}
    \caption{Overview of Rep2Text. The last-token representation obtained from the $l$-th layer of the target model $\mathcal{M}$ is projected into the embedding space of the decoding model $\mathcal{M}^\prime$ via the adapter. The projected embeddings, together with those of the system and the user prompts, are then fed into the decoding model to reconstruct the corresponding text sequence.}\label{fig:framework}
\end{figure*}

Motivated by this, we propose \textbf{Rep2Text} (Representation to Text), a framework for decoding text from the last-token representations of LLMs, as illustrated in Figure~\ref{fig:framework}. Inspired by large vision language models (LVLMs) such as LLaVA~\citep{liu2023visual}, Rep2Text trains a representation inverter that consists of a decoding language model and an adapter. The adapter maps the target representation into the decoding model's token embedding space, aligning their latent spaces.
%This mapping establishes an alignment between the latent representation space of the target model and the token embedding space of the decoding model. 
These projected embeddings are subsequently fed into the decoding LLM, enabling it to interpret them and generate text consistent with the original input sequence. By comparing the inverted text against the original text, we quantify the information retained in the last-token representation. 

Our experiments reveal that, remarkably, \emph{on held-out Wikipedia sequences of length 16, Rep2Text recovers roughly half of the original tokens from a single last-token representation, while maintaining strong semantic coherence and requiring no iterative search}. This finding directly answers our central research question and demonstrates that, although last-token representations are optimized as an information bottleneck for next-token prediction, they still retain a substantial amount of recoverable information about the input sequence.
To validate the effectiveness of our approach, we combine established quantitative metrics with LLM-as-a-judge evaluations to measure information retention at the token, structural, and semantic levels. Our results show that representations from different models exhibit varying recovery rates, revealing potential vulnerabilities in some models, while decoder family matters more than decoder scale. Further analysis shows that structural information is most prominent in early-to-middle layers, whereas semantic information becomes more pronounced in middle-to-late layers. We also find that recovery is strong for sequences shorter than 16 tokens but degrades for longer inputs. 

\section{Related Work}\label{sec:related}
\paragraph{Embedding Inversion.} The inversion is typically formulated as an optimization problem in which the attack model attempts to generate hypotheses that produce embeddings as close as possible to the target embeddings. A few works attempt to recover ground-truth sequences using sentence embeddings from BERT models.~\citet{song2020information} invert the sentence embeddings into a bag of words. Some works attempt to train an attacker model to decode the ground-truth sequence utilizing sentence embeddings and text embeddings~\citep{li2023sentence, morris2023text, huang2024transferable}. Further,~\citet{dong2025depth} extends embedding inversion to LLM's internal states at a certain layer, by learning token embeddings that can produce similar internal states. This method needs the internal states of all tokens. Thus, it does not apply to a single last-token representation. However, to fully recover the input text, these papers either rely heavily on iterative optimization or incorporate all sentence embeddings and token embeddings. \citet{nikolaou2026language} prove that LLMs map different inputs to different hidden states. They also recover the exact input from these hidden states. In contrast, we ask how much of the input a single last-token representation retains.

\paragraph{Activation Decoding.} Some work decodes activations into natural language. Recent work such as SelfIE~\citep{chen2024selfie}, and Patchscopes~\citep{ghandeharioun2024patchscopes} interpret representations through patching them into the forward pass of LLMs to decode natural language explanations. Besides, LIT~\citep{pan2024latentqa} finetunes target model to answer questions related to given activations patched within. PCD~\citep{huang2025predictive} maps activations into concept vectors then train a model to answer questions about activations. Universal Activation Verbalizer~\citep{zhao2026universal} trains a shared decoder to explain activations from different donor models. \citet{zhao2026faithful} first reconstruct the source text from an activation and then use preference optimization to reduce hallucinated descriptions. SAEExplainer~\citep{he2026saeexplainer} uses activation scores as rewards to train an explainer for SAE features. Some work explains activations by assuming that similar meanings produce similar activations. They invert an activation to find inputs that would recreate it, and use those inputs as the explanation. InverseView~\citep{huang2024inversionview} trains a decoder to sample the input distribution for a given activation. InverseScope~\citep{luo2025inversescope} explore task-specific features encoded in the input distribution. 
In contrast, we recover information solely from the last-token representation, without using the original input, partial input spans, or task-specific auxiliary hints.

\section{Rep2Text Framework}\label{sec:method}
In this section, we introduce the proposed \textit{Rep2Text} framework (see Figure~\ref{fig:framework}). Rep2Text employs a trainable adapter that bridges the target model's representation space to a decoding language model's embedding space, enabling us to systematically investigate what information is preserved in compressed last-token representations and how much of the original input can be recovered. The decoding LLM then autoregressively reconstructs the text from these projected embeddings.
%Our key insight is to leverage the learned knowledge and world models inherent in LLMs to effectively invert representations. By utilizing a separate decoding LLM, we can exploit its language understanding capabilities to reconstruct the original input from compressed token representations.

% \subsection{Problem Statement}
% Given a layer-level representation from an LLM, we aims at inversing its ground-truth sequence as much as possible to investigate to what extent we can inverse input information. 
% We term token representation at different layers of decoder-only models as representation or activation. In the following text, we will use representation and activations interchangeably.
% Concretely, give an ground-truth sequence of $n$ tokens $S=\left\langle s_1, \ldots, s_n\right\rangle$ and a \textit{target model} $\mathcal{M}$ with $L$ layers. We only focus on last-token representation, $\boldsymbol{h}^{\ell}$ denotes last-token representation of $S$ at layer $\ell \in [1, \ldots, L]$ from model $\mathcal{M}$. We propose to decode $\boldsymbol{h}^{\ell}$ into input text $\hat{S}=\left\langle \hat{s}_1, \ldots, \hat{s}_m\right\rangle$. Our goal is to study how much information is preserved in the bottleneck representation $\boldsymbol{h}^{\ell}$ by comparing the difference between original input $S$ and the decoded input $\hat{S}$.
\subsection{Problem Statement}

Given a token-level representation from an LLM, our objective is to reconstruct its original input sequence and quantify how much input information is retained. 
Throughout this work, we use \textit{representation} and \textit{activation} interchangeably to denote hidden states from decoder-only LLMs.

Formally, let $S=\langle s_1,\ldots,s_n\rangle$ be an input sequence and $\mathcal{M}$ be a target model with $L$ layers. 
For layer $\ell \in \{1,\ldots,L\}$, let $\boldsymbol{h}^{\ell}$ denote the residual-stream representation of the last token after processing $S$. 
We aim to decode $\boldsymbol{h}^{\ell}$ into an inverted sequence $\hat{S}=\langle \hat{s}_1,\ldots,\hat{s}_m\rangle$, and measure the information preserved in $\boldsymbol{h}^{\ell}$ by comparing $\hat{S}$ with the original input $S$.

\subsection{Rep2Text Inverter Design}\label{sec:inver-design}

To invert the representation, we propose an \underline{\emph{inverter}} architecture inspired by the design of typical large vision-language models such as LLaVA. The inverter consists of two key components: (1) a trainable \emph{adapter} that projects the target model's internal representation into the input token embedding space of the decoding language model, and (2) a \emph{decoding language model} that generates the inverted text from these projected embeddings.

Specifically, we introduce a decoding model $\mathcal{M}^\prime$ that can either be a copy of the target model $\mathcal{M}$ or a different LLM. To bridge the representation space of $\mathcal{M}$ and the embedding space of $\mathcal{M}^\prime$, we train an adapter to project the token representation $\boldsymbol{h}^\ell \in \mathbb{R}^d$ from the target model $\mathcal{M}$ into the token embedding space of the decoding model $\mathcal{M}^\prime$. The adapter is implemented as a two-layer MLP with gated skip connection with optional projection, defined as:
% \begin{equation}\label{eq:mlp}
% \hspace{-0.2cm}\begin{aligned}
%     &\boldsymbol{h}_1 = \mathrm{GELU}(\boldsymbol{W}_1 \cdot \mathrm{LN}(\boldsymbol{h}_\ell) + \boldsymbol{b}_1),\\
%     &\boldsymbol{h}_2 = \boldsymbol{W}_2 \cdot \boldsymbol{h}_1 + \boldsymbol{b}_2,\\
%     &[\tilde{\boldsymbol{x}}_1; \ldots; \tilde{\boldsymbol{x}}_k] = \mathrm{reshape}(\boldsymbol{h}_2, k, d^\prime),\\
%     &\boldsymbol{x}_i = \mathrm{LN}(\boldsymbol{W}_s \cdot \boldsymbol{h}_\ell + g_i \cdot \tilde{\boldsymbol{x}}_i), \quad i = 1,\ldots,k,\\
%     &\boldsymbol{X}_e = [\boldsymbol{x}_1; \ldots; \boldsymbol{x}_k],
% \end{aligned}
% \end{equation}
\begin{equation}\label{eq:mlp}
\begin{aligned}
    &\boldsymbol{h}_1 = \mathrm{GELU}(\boldsymbol{W}_1 \cdot \mathrm{LN}(\boldsymbol{h}_\ell) + \boldsymbol{b}_1), \quad
    \boldsymbol{h}_2 = \boldsymbol{W}_2 \cdot \boldsymbol{h}_1 + \boldsymbol{b}_2, \quad
    [\tilde{\boldsymbol{x}}_1; \ldots; \tilde{\boldsymbol{x}}_k] = \mathrm{reshape}(\boldsymbol{h}_2, k, d^\prime),\\
    &\boldsymbol{x}_i = \mathrm{LN}(\boldsymbol{W}_s \cdot \boldsymbol{h}_\ell + g_i \cdot \tilde{\boldsymbol{x}}_i), \quad i = 1,\ldots,k, \quad
    \boldsymbol{X}_e = [\boldsymbol{x}_1; \ldots; \boldsymbol{x}_k].
\end{aligned}
\end{equation}
\noindent where 
% for the first MLP layer we use dropout $\mathrm{Drop}(\cdot,p)$  with probability $p$. 
$\mathrm{LN}(\cdot)$ and $\mathrm{GELU}(\cdot)$ represent the norm layer and activation function respectively. 
$\boldsymbol{W}_1\in \mathbb{R}^{d \times d^{\mathrm{hid}}}$ and $\boldsymbol{W}_2\in \mathbb{R}^{d^{\mathrm{hid}} \times k\cdot d^{\prime}}$ refer to linear transformations in the first and second layers respectively, where $d$ and $d^\prime$ represent the hidden dimensions of the target model and decoding model. Note that we set $d^{\mathrm{hid}} = f\cdot d$, where $f$ is an expansion factor. 
%$\boldsymbol{W}_s \in \mathbb{R}^{d\times d^\prime}$ denotes the transformation matrix of the skip connection. When $d = d^\prime$, it is an identity matrix. 
$\boldsymbol{W}_s \in \mathbb{R}^{d \times d'}$ denotes the transformation matrix of 
the skip connection. When $d = d'$, $\boldsymbol{W}_s$ is an identity matrix enabling 
a true residual connection; when $d \neq d'$, $\boldsymbol{W}_s$ serves as a learned 
projection matrix to match dimensions. $\boldsymbol{h}_2\in\mathbb{R}^{k \cdot d^\prime}$ is reshaped into $(k, d^\prime)$, which can be regarded as $k$ token embeddings. Each token embedding is constructed with a gated combination of the skip path and the MLP-transformed path to preserve the representation information as much as possible. The projected token embedding can be denoted as $\boldsymbol{X}_e = [\boldsymbol{x}_1; \cdots ;\boldsymbol{x}_k]$, where the number $k$ of projected tokens is a hyperparameter, and $g_i$ is a learnable scalar gate for the i-th projected token embedding, and the skip-path vector $\boldsymbol{W}_s \cdot \boldsymbol{h}^\ell$ is broadcast to each projected token position.

For each representation, the projected token embedding $\boldsymbol{X}_e$ is combined with system prompt embedding $\boldsymbol{X}_{\mathrm{sys}}$ and user prompt embedding $\boldsymbol{X}_u$. The combined sequence $[\boldsymbol{X}_e; \boldsymbol{X}_{\mathrm{sys}}; \boldsymbol{X}_u]$ is fed into the first layer of decoding model $\mathcal{M}^\prime$ (after its embedding layer), bypassing the embedding layer. The decoding model then autoregressively generates the inverted text $\hat{S}$.

\subsection{Rep2Text Inverter Training}\label{sec:training}

For an output sequence of length $T$, at each decoding step $t$, the inverter predicts its probability conditioned on the projected embeddings, prompt embeddings, and all previously generated tokens. The joint probability of inverted sequence $\hat{S}$ is:
% \begin{equation}\label{eq:regressive}
% \begin{aligned}
% p\left(\hat{S} \mid\right.&\left. \boldsymbol{X}_e, \boldsymbol{X}_{\mathrm{sys}},\boldsymbol{X}_u\right)\\
% &=\prod_{t=1}^T p_\theta\left(\hat{s}_t \mid \boldsymbol{X}_e, \boldsymbol{X}_{\mathrm{sys}}, \boldsymbol{X}_u, \hat{S}_{<t}\right),
% \end{aligned}
% \end{equation}
\begin{equation}\label{eq:regressive}
p\left(\hat{S} \mid \boldsymbol{X}_e, \boldsymbol{X}_{\mathrm{sys}},\boldsymbol{X}_u\right) = \prod_{t=1}^T p_\theta\left(\hat{s}_t \mid \boldsymbol{X}_e, \boldsymbol{X}_{\mathrm{sys}}, \boldsymbol{X}_u, \hat{S}_{<t}\right)
\end{equation}
where $\hat{S}_{<t}$ are the inverted tokens generated before step $t$, and $\theta$ denotes the trainable parameters. We consider two training schemes: (1) \textit{adapter-only tuning}, where only the adapter parameters are optimized; (2) \textit{full fine-tuning}, where the adapter is first fine-tuned independently and then the adapter continues to be fine-tuned while the decoding model is updated via LoRA~\citep{hu2022lora}. Accordingly, $\theta$ refers to the trainable parameters under the chosen scheme. 

During training, we employ teacher forcing: the generated prefix $\hat{S}_{<t}$ is replaced with the ground-truth prefix $S_{<t}$, and the model is optimized to maximize the log-likelihood of the ground-truth token $s_t$ at each step. To stabilize training, we use label smoothing to soften the one-hot target distribution. The ground-truth token vocabulary distribution is denoted as
% During training, we employ teacher forcing to maximize the log-likelihood of ground-truth tokens $s_t$ at each step. To stabilize training, we utilize label smoothing to soften the one-hot target distribution. The ground-truth token vocabulary distribution is denoted as
\begin{equation}\label{eq:label-smooth}
q_t(v_i)=(1-\epsilon) \mathbf{1}\left[v_i=s_t\right]+\frac{\epsilon}{|V|},
\end{equation}
where $\mathbf{1}(\cdot)$ is an indicator function that equals 1 if the condition holds and 0 otherwise, and $\epsilon$ is the label smoothing factor, set to be 0.075. The training objective is the smoothed cross-entropy loss:
% \begin{small}
% \begin{equation}
% \begin{aligned}
% &\mathcal{L}_t = -\sum_{i=1}^{|V|} q_t(v_i)\, \log p_\theta\!\left(v_i \mid \mathbf{X}_e, \mathbf{X}_{\mathrm{sys}}, \mathbf{X}_u, {S}_{<t}\right), \\
% &\mathcal{L}_{\mathrm{LS}} = \frac{1}{T} \sum_{t=1}^T \mathcal{L}_t.
% \end{aligned}
% \end{equation}
\begin{equation}
\mathcal{L}_t = -\sum_{i=1}^{|V|} q_t(v_i)\, \log p_\theta\!\left(v_i \mid \mathbf{X}_e, \mathbf{X}_{\mathrm{sys}}, \mathbf{X}_u, {S}_{<t}\right), \qquad \mathcal{L}_{\mathrm{LS}} = \frac{1}{T} \sum_{t=1}^T \mathcal{L}_t.
\end{equation}
% \end{small}
This training objective optimizes the adapter (and optionally the decoding model via LoRA) to minimize the prediction error across all token positions 
in the inverted sequence. A label smoothing term is incorporated to prevent overconfidence in token predictions, thereby improving generalization 
to unseen representations. Through this training process, the adapter learns 
to effectively map the compressed last-token representation from the target 
model's latent space into the decoding model's token embedding space, enabling 
the reconstruction of the original input sequence.

% As our objective is decoding the exact input text, we use greedy decoding to prevent models from creative generation such as rephrasing or 

\section{Experiments}\label{sec:exp}
In this section, we first examine the training strategy and data scale ($\S$\ref{sec:training-strategy}). 
Next, we evaluate Rep2Text across target and decoding model combinations ($\S$\ref{sec:other-model}). 
We then conduct a representation--text correspondence analysis to demonstrate that inversion relies on primarily representations rather than decoder language priors ($\S$\ref{sec:random}). 
To analyze the information bottleneck in last-token representations, we study inversion performance across varying input lengths ($\S$\ref{sec:token-length}) and different target-model layers ($\S$\ref{sec:layerwise}). 
Finally, we compare Rep2Text with the baseline method on both in-distribution and out-of-distribution datasets to evaluate its generalizability ($\S$\ref{sec:case}).

\subsection{Experimental Setup}\label{sec:setup}

\paragraph{Datasets.} We train adapters on passages randomly truncated from Wikipedia articles in \textit{The Pile}~\citep{gao2020pile}. Each passage contains $n$ non-overlapping tokens, where $n \in \{8, 16, 32, 64, 128, 256\}$ depending on the experimental setting. Each training example pairs the last-token representation from a fixed layer of the target model with its corresponding ground-truth input sequence. We use 640K sequences for adapter-only tuning and add 960K sequences during full fine-tuning. For evaluation, we randomly sample 1,000 sequences as the test set and assess the inverted outputs using both quantitative metrics and LLM-as-a-judge evaluations.

% \paragraph{Models} Our paper use Llama 3 series models as the decoding models. We mainly use Llama-3.1-8B~\footnote{https://huggingface.co/meta-llama/Llama-3.1-8B} as an inverter to decode its own representations in $\S$\ref{sec:token-length}, $\S$\ref{sec:layerwise}, $\S$\ref{sec:other-model}, and $\S$\ref{sec:ood}. Besides, in $\S$\ref{sec:other-model}, we use Gemma-7B~\footnote{https://huggingface.co/google/gemma-7b} and Mistral-7B-v0.1~\footnote{https://huggingface.co/mistralai/Mistral-7B-v0.1} as target models and Llama-3.1-8B as the inverter to study the feasibility and effectiveness of using another model as the decoding model. To study the scaling law of representation inversion, we adopt Llama-3.2-3B~\footnote{https://huggingface.co/meta-llama/Llama-3.2-3B} and Llama-3.1-70B~\footnote{https://huggingface.co/meta-llama/Llama-3.1-70B} as an comparison to illustrate if larger model improve the performance of inverters.
\paragraph{Models.} We evaluate multiple target--decoder model combinations. The decoding models include Llama-3.2-3B~\citep{llama3-2-3b}, Llama-3.1-8B~\citep{llama3-1-8b}, Qwen-2.5-14B~\citep{qwen2.5}, Qwen-2.5-32B~\citep{qwen2.5}, and Qwen-2.5-72B~\citep{qwen2.5}. The target models include Llama-3.2-3B, Llama-3.1-8B, Mistral-7B-v0.1~\citep{mistral-7b-v0.1}, and Gemma-7B~\citep{gemma-7b}. To study cross-model inversion, we fix Llama-3.1-8B as the decoder and evaluate it across all target models. To analyze scaling behavior, we invert representations from Mistral-7B-v0.1 using decoders of increasing size, including Llama-3.2-3B, Llama-3.1-8B, Qwen-2.5-14B, Qwen-2.5-32B, and Qwen-2.5-72B.

\begin{figure}[!t]
    \centering    \includegraphics[width=\linewidth]{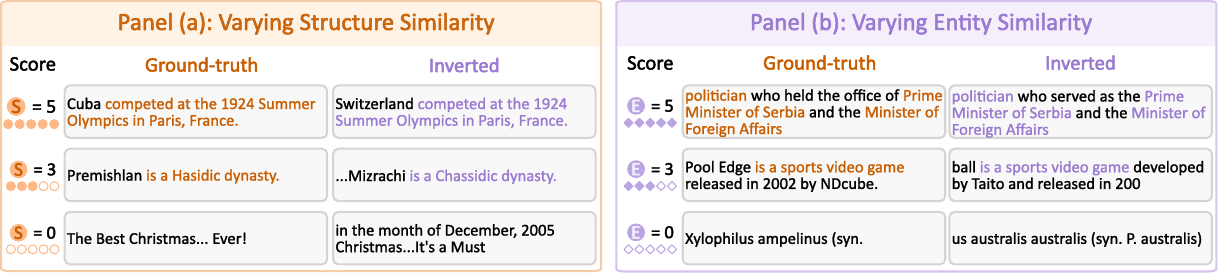}
    \caption{Examples of structural and entity preservation scores. S and E denote structural and entity similarity, respectively, rated on a 0--5 scale using an LLM-as-a-judge.}
    \label{fig:example}
\end{figure}

\paragraph{Implementation Details.}
Our main experiments use sequences with $n=16$ tokens. 
For models up to 14B, adapter-only tuning on two A100 GPUs takes 7 hours, while full fine-tuning requires additional 12 hours. 
We use a peak adapter learning rate of $10^{-3}$, except for Qwen models. 
For full fine-tuning, we use learning rates of $5\times10^{-4}$ for the adapter and $2\times10^{-4}$ for LoRA parameters. 
Models are trained for 3 epochs, except Qwen models, which are trained for 5 epochs; longer training yields only marginal gains. 
Additional hyperparameters are provided in Appendix~\ref{app:training}. 
Unless otherwise specified, all experiments fine-tune only the adapter while keeping the decoder frozen.

\paragraph{Evaluation Measurements.}\label{sec:evaluation}
We evaluate inverted sequences along three dimensions: token-level recovery, structural and entity preservation, and semantic similarity. We use \textit{ROUGE-1}, \textit{ROUGE-2}, and \textit{ROUGE-L} to measure token-level recovery. We use GPT-4.1-mini to rate structural and entity preservation on a 0--5 scale, which is normalized to 0--1. As shown in Figure~\ref{fig:example}, these scores assess whether the inverted text preserves the grammatical structure and key entities of the ground truth. We quantify semantic similarity using \textit{BERTScore} and LLM-based topic relevance. Detailed metric definitions are provided in Appendix~\ref{app:metric}.

\subsection{Training Strategy and Data Scale}\label{sec:training-strategy}
\begin{figure}[!t]
    \centering
    \includegraphics[width=\linewidth]{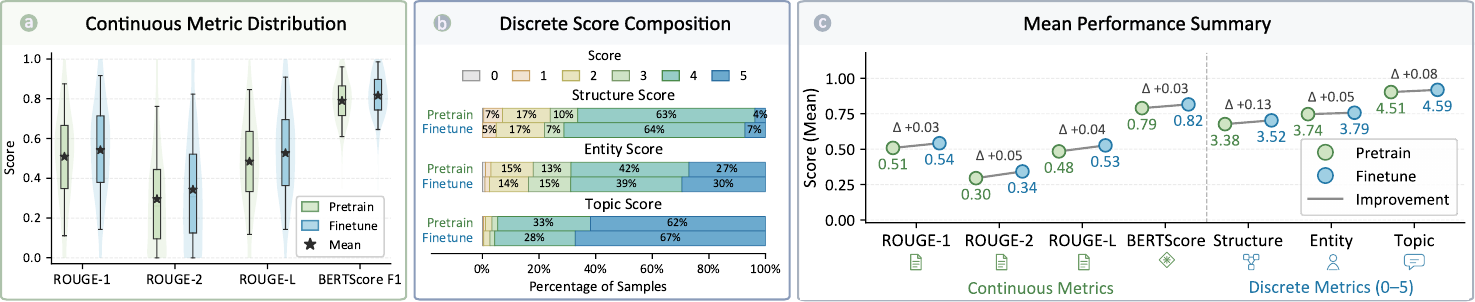}
    \caption{Adapter-only tuning versus full fine-tuning. Decoder fine-tuning improves token recovery but yields limited gains on semantic and LLM-judge metrics.}
    \label{fig:ft}
\end{figure}

We first examine two practical factors in training Rep2Text: whether the decoder needs to be updated, and how much adapter training data is required.

\paragraph{Effect of Decoder Fine-tuning.} We compare \textit{adapter-only tuning}, where the decoder is frozen, with \textit{full fine-tuning}, where the trained adapter is further optimized together with LoRA parameters on the decoder. 
Figures~\ref{fig:ft}(a) and~\ref{fig:ft}(b) show that full fine-tuning improves inversion performance across all metrics, but the gains are modest. 
As summarized in Figure~\ref{fig:ft}(c), full fine-tuning increases ROUGE-1, ROUGE-2, and ROUGE-L by $0.03$, $0.05$, and $0.04$, respectively, with similarly limited gains on semantic and LLM-judge metrics. 
The score distributions further show that full fine-tuning mainly shifts samples toward higher token-overlap scores, while semantic metrics remain largely similar. 
These results suggest that representation-to-embedding alignment is primarily learned by the adapter. 
Therefore, all following experiments use adapter-only tuning and keep the decoder frozen.

\begin{table}[t]
\centering
\small
\caption{
Performance comparison of representation inversion across target and decoding model combinations.
Group (A) uses the best layer of each target model, which is the layer with the highest ROUGE-1 in the layer-wise sweep.
Groups (B) and (C) fix layer 10, so that all decoders invert the same representations.
Bold marks the best score in each column within each group.
}\label{tab:model}
\setlength{\tabcolsep}{2pt}
\resizebox{\linewidth}{!}{
\begin{tabular}{llcccccccc}
\toprule
\textbf{Target Model} & \textbf{Decoding Model} & \textbf{Layer} & \textbf{ROUGE-1} & \textbf{ROUGE-2} & \textbf{ROUGE-L} & \textbf{BERTScore} & \textbf{Structure} & \textbf{Entity} & \textbf{Topic} \\
& & & \textbf{(0-1)$\uparrow$} & \textbf{(0-1)$\uparrow$} & \textbf{(0-1)$\uparrow$} & \textbf{(0-1)$\uparrow$} & \textbf{(0-1)$\uparrow$} & \textbf{(0-1)$\uparrow$} & \textbf{(0-1)$\uparrow$} \\
\midrule
\multicolumn{10}{l}{\textit{(A) Fixed Llama-3.1-8B decoder, best layer per target: target invertibility}} \\
Gemma-7B & Llama-3.1-8B & L26
& \textbf{0.64}\std{0.24} & \textbf{0.42}\std{0.30} & \textbf{0.61}\std{0.25} & 0.80\std{0.14} & \textbf{0.71}\std{0.22} & 0.64\std{0.27} & 0.86\std{0.22} \\
Mistral-7B-v0.1 & Llama-3.1-8B & L15
& 0.55\std{0.25} & 0.35\std{0.28} & 0.53\std{0.25} & \textbf{0.81}\std{0.13} & 0.70\std{0.23} & \textbf{0.75}\std{0.25} & \textbf{0.92}\std{0.17} \\
Llama-3.1-8B & Llama-3.1-8B & L10
& 0.48\std{0.23} & 0.28\std{0.24} & 0.47\std{0.22} & 0.78\std{0.11} & 0.66\std{0.22} & 0.74\std{0.23} & 0.91\std{0.14} \\
Llama-3.2-3B & Llama-3.1-8B & L9
& 0.46\std{0.22} & 0.25\std{0.22} & 0.44\std{0.21} & 0.76\std{0.11} & 0.64\std{0.22} & 0.72\std{0.24} & 0.89\std{0.16} \\
\midrule
\multicolumn{10}{l}{\textit{(B) Self vs.\ cross-model decoding (layer 10)}} \\
Llama-3.2-3B & Llama-3.2-3B & L10
& \textbf{0.46}\std{0.22} & \textbf{0.26}\std{0.23} & \textbf{0.45}\std{0.21} & \textbf{0.76}\std{0.11} & \textbf{0.64}\std{0.22} & 0.69\std{0.24} & \textbf{0.88}\std{0.16} \\
Llama-3.2-3B & Llama-3.1-8B & L10
& 0.45\std{0.22} & 0.25\std{0.22} & 0.43\std{0.22} & \textbf{0.76}\std{0.11} & \textbf{0.64}\std{0.22} & \textbf{0.72}\std{0.23} & \textbf{0.88}\std{0.16} \\
\midrule
\multicolumn{10}{l}{\textit{(C) Decoder scale and family (Mistral-7B-v0.1, layer 10)}} \\
Mistral-7B-v0.1 & Llama-3.2-3B & L10
& \textbf{0.52}\std{0.23} & \textbf{0.32}\std{0.26} & 0.50\std{0.23} & 0.80\std{0.12} & 0.70\std{0.21} & 0.73\std{0.25} & 0.90\std{0.17} \\
Mistral-7B-v0.1 & Llama-3.1-8B & L10
& \textbf{0.52}\std{0.23} & \textbf{0.32}\std{0.26} & \textbf{0.51}\std{0.23} & \textbf{0.81}\std{0.11} & \textbf{0.71}\std{0.21} & \textbf{0.75}\std{0.23} & 0.90\std{0.18} \\
Mistral-7B-v0.1 & Qwen-2.5-14B & L10
& 0.48\std{0.21} & 0.27\std{0.23} & 0.47\std{0.21} & 0.78\std{0.11} & 0.65\std{0.23} & 0.67\std{0.25} & 0.89\std{0.18} \\
Mistral-7B-v0.1 & Qwen-2.5-32B & L10
& 0.47\std{0.21} & 0.25\std{0.23} & 0.45\std{0.21} & 0.76\std{0.12} & 0.65\std{0.24} & 0.65\std{0.26} & 0.88\std{0.19} \\
Mistral-7B-v0.1 & Qwen-2.5-72B & L10
& 0.51\std{0.23} & 0.30\std{0.25} & 0.49\std{0.23} & 0.78\std{0.12} & 0.68\std{0.23} & 0.74\std{0.25} & \textbf{0.91}\std{0.17} \\
\bottomrule
\end{tabular}
}
\end{table}

\paragraph{Effect of Training Dataset Size.} We additionally study how the amount of adapter training data affects inversion performance. As shown in Appendix~\ref{app:dataset}, increasing the dataset size from 10K to 640K leads to monotonic improvements across all eight metrics. However, the improvement from 320K to 640K shows diminishing returns around the 640K setting, making it a reasonable compute-performance trade-off for our main experiments.

\subsection{Cross-Model Representation Inversion}\label{sec:other-model}
In this section, we study whether a single last-token representation can be inverted across different target and decoding models. We aim to answer four questions: 
(i) how much information can be recovered from a single last-token representation, 
(ii) whether some target models expose more recoverable information than others, 
(iii) whether cross-model decoding is as effective as self-model decoding, and 
(iv) whether larger decoding models improve inversion quality. We evaluate 16-token Wikipedia sequences and report results in Table~\ref{tab:model}. For (ii), we use the best layer of each target model, selected by ROUGE-1 in the layer-wise sweep ($\S$\ref{sec:layerwise}). For (iii) and (iv), we fix layer 10, so that all decoders invert the same representations. Across all aforementioned experiments, only the adapter is trained and the decoding model is kept frozen.

\paragraph{Single-token representations retain substantial recoverable information.}
Across target--decoder combinations, Rep2Text recovers substantial information from a single last-token representation. 
ROUGE-1 ranges from 0.45 to 0.64, suggesting that roughly half of the unigrams are recovered in most settings, while ROUGE-2 remains at 0.25--0.42, indicating non-trivial local phrase recovery. 
The inverted sequences also preserve strong semantic similarity, with BERTScore between 0.76 and 0.81 and topic preservation between 0.86 and 0.92. 
These results show that the last-token representation is a strong but not opaque information bottleneck, retaining both lexical fragments and high-level meaning.

\paragraph{Target models differ in representation invertibility.}
We  fix Llama-3.1-8B as the decoder and compare representations from Gemma-7B, Mistral-7B-v0.1, Llama-3.1-8B, and Llama-3.2-3B, each at its best layer. 
Gemma-7B achieves the strongest lexical and structural recovery at layer 26, with 0.64 ROUGE-1, 0.42 ROUGE-2, 0.61 ROUGE-L, and 0.71 structure score. 
Mistral-7B-v0.1 achieves the strongest semantic recovery at layer 15, with 0.81 BERTScore, 0.75 entity score, and 0.92 topic score. 
By contrast, Llama-family targets are less recoverable, with Llama-3.1-8B and Llama-3.2-3B achieving 0.48 and 0.46 ROUGE-1, respectively. 
This suggests that sharing the same model family with the decoder does not necessarily make inversion easier; instead, recoverable input-specific information in last-token representations varies across model families. 
The best layer also varies across model families ($\S$\ref{sec:layerwise}).

\paragraph{Cross-model decoding is robust.}
We then compare self-model and cross-model decoding at layer 10. 
Using Llama-3.2-3B as both the target and decoder achieves 0.46 ROUGE-1, 0.26 ROUGE-2, and 0.45 ROUGE-L, comparable to decoding the same target representations with Llama-3.1-8B. 
Similarly, for Mistral-7B-v0.1 representations, Llama-3.1-8B and Llama-3.2-3B perform nearly identically, both reaching 0.52 ROUGE-1 and 0.32 ROUGE-2. 
These results show that successful inversion does not require the decoder to match the target model. 
Instead, the adapter can bridge the target representation space and decoder embedding space, suggesting that recoverable input information is partially aligned across different LLMs.

\noindent\textbf{Decoder family matters more than decoder scale.}
Finally, we examine whether increasing decoder size improves inversion by replacing the Llama decoder with Qwen-2.5-14B, Qwen-2.5-32B, and Qwen-2.5-72B. Despite having more than twice as many parameters, Qwen-2.5-32B does not improve over Qwen-2.5-14B. Instead, performance slightly decreases from 0.48 to 0.47 in ROUGE-1, from 0.27 to 0.25 in ROUGE-2, and from 0.78 to 0.76 in BERTScore. Qwen-2.5-72B improves over Qwen-2.5-32B, but it still underperforms the smaller Llama decoders on most metrics when inverting Mistral-7B-v0.1 representations. These results suggest that cross-model decoding is robust to decoder scale but sensitive to decoder family. 
% Overall, Table~\ref{tab:model} shows effective cross-model inversion, with recoverability depending more on the target model than decoder scale.

\subsection{Representation-Text Correspondence Analysis}\label{sec:random}
To test whether Rep2Text relies on representation-specific information rather than decoder language priors, we compare three settings: \textit{Baseline}, which uses correctly paired representations and original inputs; \textit{Reverse}, which reconstructs word-level reversed and retokenized inputs; and \textit{Random}, which pairs each representation with an unrelated target sequence and breaks the alignment.
Random serves as a negative control. Any performance under Random comes from decoder language priors alone.

Figure~\ref{fig:random} shows that breaking the representation--text correspondence leads to a clear performance collapse. 
The \textit{Random} setting performs substantially worse than the correctly paired \textit{Baseline} setting across loss, lexical overlap, and semantic similarity, indicating that Rep2Text cannot recover meaningful text from unrelated representation--text pairs and does not rely solely on decoder language priors. 
In contrast, \textit{Reverse} remains partially learnable despite violating the natural left-to-right word order favored by LLMs. 
This suggests that the last-token representation still contains recoverable input-specific information under a deterministic word-order transformation. 
Detailed experimental settings and quantitative results are provided in Appendix~\ref{app:random}.

\begin{figure*}[t]
    \centering
    \begin{minipage}[t]{0.485\linewidth}
    \centering
    \subfloat[]{
        \includegraphics[width=0.47\linewidth]{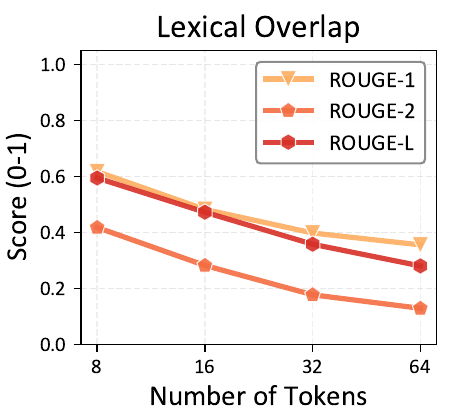}
        \label{fig:token-lexcial}
    }
    \hfill
    \subfloat[]{
        \includegraphics[width=0.47\linewidth]{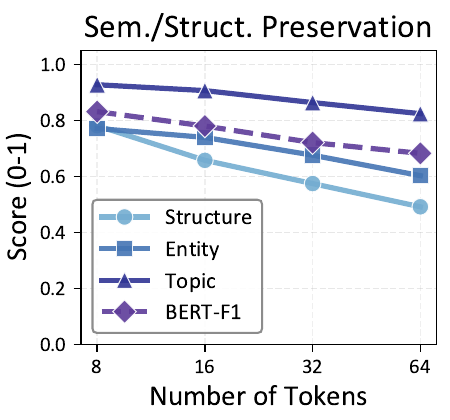}
        \label{fig:token-semantic}
    }
    \caption{Performance comparison under different input sequence lengths.}
    \label{fig:token}
    \end{minipage}
    \hfill
    \begin{minipage}[t]{0.485\linewidth}
    \centering
    \subfloat[]{
        \includegraphics[width=0.47\linewidth]{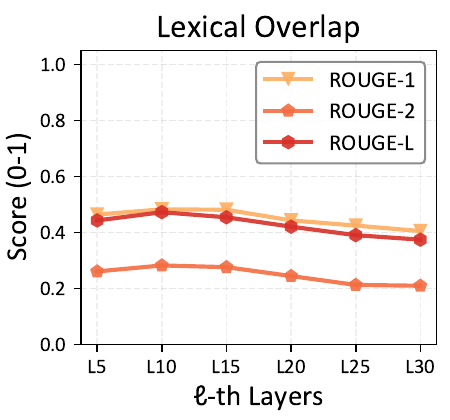}\label{fig:layer-lexcial}}
    \hfill
    \subfloat[]{
        \includegraphics[width=0.47\linewidth]{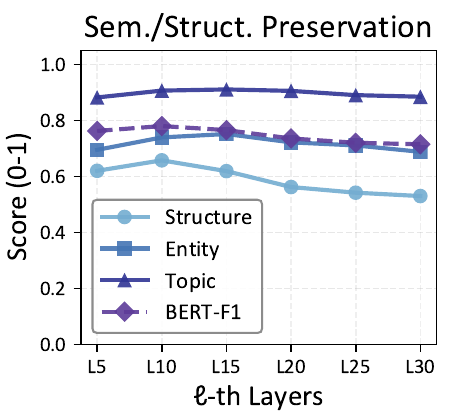}\label{fig:layer-semantic}}
    \caption{Performance comparison with layerwise representation inversions.}\label{fig:layer}
    \end{minipage}
\end{figure*}

\subsection{Effects of Sequence Length}\label{sec:token-length}
We next examine how input length affects recoverability from a single last-token representation. 
Because this representation summarizes all preceding tokens, longer inputs should create a stronger information bottleneck. 
We train separate adapters to invert layer-10 representations from 8-, 16-, 32-, and 64-token Wikipedia sequences, using Llama-3.1-8B as both the target and decoding model based on Section~\ref{sec:other-model}.

Figure~\ref{fig:token} shows that lexical recovery consistently decreases with input length. 
ROUGE-1 drops from about 0.60 for 8-token inputs to about 0.36 for 64-token inputs, with similar declines in ROUGE-2 and ROUGE-L. 
Structural preservation also degrades, suggesting that longer contexts make detailed wording and grammar harder to recover from a single representation. 
In contrast, semantic metrics decline more slowly: BERTScore, entity preservation, and topic preservation remain relatively high for longer inputs. 
Overall, a single last-token representation preserves the general topic of longer contexts better than exact wording or sentence structure. 
We further test 128- and 256-token inputs with Llama-3.2-3B as both the target and decoder.
ROUGE-1 drops from 0.46 at 16 tokens to 0.30 at 128 and 256 tokens.
In contrast, BERTScore stays at 0.63 and the topic score stays at or above 0.68, as detailed in Appendix~\ref{app:long}.

\begin{wrapfigure}{r}{0.45\linewidth}
    \vspace{-1.2em}
    \centering
    \includegraphics[width=\linewidth]{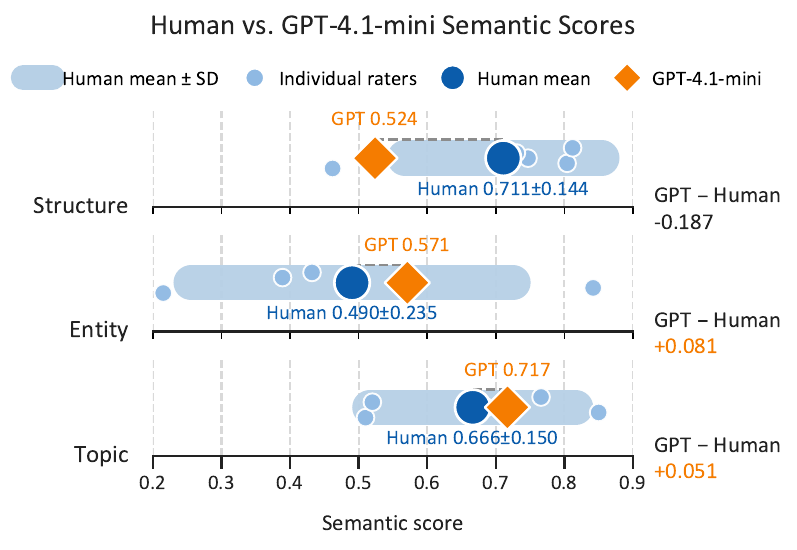}
    \caption{Human evaluation sanity check for GPT-4.1-mini semantic scores.}
    \label{fig:human_sanity}
    \vspace{-2.0em}
\end{wrapfigure}

\subsection{Layer-wise Invertibility}\label{sec:layerwise}
Prior work shows that middle-to-deep layers capture high-level semantics more effectively~\citep{jin2024exploring, campbell2023localizing}. 
We therefore ask which layer contains the most recoverable input information. 
To answer this, we train separate adapters on last-token representations from layers 5, 10, 15, 20, 25, and 30 of Llama-3.1-8B, using 16-token Wikipedia sequences.

Figure~\ref{fig:layer} shows that recoverability is strongest around L10--L15, with different metrics peaking at slightly different depths. 
ROUGE-L, structure score, and BERTScore peak around layer 10, suggesting that sentence structure, local phrase organization, and semantic similarity are already well captured by early-to-middle layers. 
ROUGE-1 is nearly identical at layers 10 and 15, while entity recovery peaks at layer 15, indicating that entity-level details may require deeper processing. 
Topic preservation remains stable from layers 10 to 20, suggesting that high-level semantic information persists across middle-to-late layers. 
Appendix~\ref{app:layer} reports layer-wise sweeps for the other three target models. The best layer depends on the model architecture.

\begin{table*}[t]
\centering
\setlength{\tabcolsep}{3pt}
\caption{Inversion comparison between Vec2Text and our method with input sequences of no more than 32 tokens.}\label{tab:baseline}
\scalebox{0.77}{\begin{tabular}{m{2.5cm}lm{3.5cm}ccccccccc}
\toprule
\textbf{Target model/} & \textbf{Data} & \textbf{Inversion Model} & \textbf{R1} & \textbf{R2} & \textbf{RL} & \textbf{Token F1} & \textbf{BLEU} & \textbf{BS} & \textbf{SS} & \textbf{ES} & \textbf{TS} \\
\textbf{decoding model} & & & \textbf{(0-1)$\uparrow$} & \textbf{(0-1)$\uparrow$} & \textbf{(0-1)$\uparrow$} & \textbf{(0-1)$\uparrow$} & \textbf{(0-1)$\uparrow$} & \textbf{(0-1)$\uparrow$} & \textbf{(0-1)$\uparrow$} &\textbf{(0-1)$\uparrow$} & \textbf{(0-1)$\uparrow$} \\
\midrule
\multirow{6}{*}{\parbox{2.4cm}{\centering Mistral-7B-v0.1/ \\ Qwen-2.5-14B}} & \multirow{3}{*}{Wiki} & \cellcolor{qblue} Rep2Text (Ours) & \cellcolor{qblue} 0.41 & \cellcolor{qblue} 0.19 & \cellcolor{qblue} 0.38 & \cellcolor{qblue} 0.42 & \cellcolor{qblue} 0.15 & \cellcolor{qblue} 0.73 & \cellcolor{qblue} 0.57 & \cellcolor{qblue} 0.66 & \cellcolor{qblue} 0.86 \\
& & Vec2Text Base &  0.38 & 0.17 & 0.35 & 0.31 & 0.15 & 0.71 & 0.53 & 0.53 & 0.80 \\
& & * + Corrector (50 Steps) &  0.38 & 0.17 & 0.35 & 0.31 & 0.15 & 0.71 & 0.53 & 0.55 & 0.79 \\
\cmidrule{2-12}
& \multirow{3}{*}{Clinical} & \cellcolor{qblue} Rep2Text (Ours) & \cellcolor{qblue} 0.37 & \cellcolor{qblue} 0.15 & \cellcolor{qblue} 0.34 & \cellcolor{qblue} 0.26 & \cellcolor{qblue} 0.10 & \cellcolor{qblue} 0.74 & \cellcolor{qblue} 0.59 & \cellcolor{qblue} 0.48 & \cellcolor{qblue} 0.64 \\
& & Vec2Text Base & 0.14 & 0.02 & 0.13 & 0.13 & 0.04 & 0.53 & 0.17 & 0.16 & 0.21 \\
& & {* + Corrector (50 Steps)} & 0.13 & 0.01  & 0.12  & 0.11 & 0.03 & 0.52 & 0.14 & 0.13 & 0.20 \\
\midrule
\multirow{6}{*}{\parbox{2.4cm}{\centering Llama-3.1-8B/ \\Qwen-2.5-14B}} & \multirow{3}{*}{Wiki} & \cellcolor{qblue} Rep2Text (Ours) & \cellcolor{qblue} 0.36 & \cellcolor{qblue} 0.15 & \cellcolor{qblue} 0.32 & \cellcolor{qblue} 0.39 & \cellcolor{qblue} 0.12 & \cellcolor{qblue} 0.70 & \cellcolor{qblue} 0.53 & \cellcolor{qblue} 0.63 & \cellcolor{qblue} 0.85 \\
& & Vec2Text Base & 0.37 & 0.16 & 0.34 & 0.29 & 0.14 & 0.70 & 0.54 & 0.51 & 0.78 \\
& & * + Corrector (50 Steps) & 0.35 & 0.15 & 0.31 & 0.27 & 0.12 & 0.68  & 0.50 & 0.49 & 0.76 \\
\cmidrule{2-12}
& \multirow{3}{*}{Clinical} & \cellcolor{qblue} Rep2Text (Ours) & \cellcolor{qblue} 0.34 & \cellcolor{qblue} 0.12 & \cellcolor{qblue} 0.31 & \cellcolor{qblue} 0.24 & \cellcolor{qblue} 0.08 & \cellcolor{qblue} 0.71 & \cellcolor{qblue} 0.50 & \cellcolor{qblue} 0.47 & \cellcolor{qblue} 0.56 \\
& & Vec2Text Base & 0.13 & 0.01 & 0.12 & 0.12 & 0.03 & 0.55 & 0.23 & 0.18 & 0.25 \\
& & {* + Corrector (50 Steps)} & 0.12 & 0.01 & 0.11 & 0.11 & 0.03 & 0.51 & 0.17 & 0.14 & 0.22  \\
\bottomrule
\multicolumn{12}{l}{\small * denotes Vec2Text}\\
\end{tabular}}
\end{table*}

\begin{table*}[t]
    \centering
    \setlength{\tabcolsep}{3pt}
    \caption{Inverted examples of the three methods on the same inputs. Bold marks the best Token F1 in each group.}
    \label{tab:clinical}
    \scalebox{0.75}{\begin{tabular}{m{3.5cm}m{6cm}m{5.5cm}cc}
    \toprule
    \textbf{Method} & \textbf{Ground-truth Sequence} & \textbf{Inverted Sequence} & \textbf{Token F1} & \textbf{BLEU} \\
    \midrule
    \multicolumn{5}{l}{\textit{In-distribution, Llama-3.1-8B target}} \\
    \midrule
    \rowcolor{qblue} Ours
      & Phil LaMarr\newline Phillip LaMarr (born January 24, 1967) is an American actor, voice actor, comedian and writer.
      & Phil LaMarr\newline Philip LaMarr (born January 24, 1967) is an American actor, voice actor, comedian and writer.
      & \textbf{0.94} & 0.89 \\
    Vec2Text Base
      &
      & David A. Smith David A. Smith (born December 16, 1964) is an American actor, voice actor, writer, and voice actor.
      & 0.43 & 0.32 \\
    Vec2Text + Corrector (50 Steps)
      &
      & David A. Smith David A. Smith (born August 24, 1964) is an American actor, voice actor, writer, and voice actor.
      & 0.49 & 0.33 \\
    \midrule
    \multicolumn{5}{l}{\textit{In-distribution, Mistral-7B-v0.1 target}} \\
    \midrule
    \rowcolor{qblue} Ours
      & List of submissions to the 37th Academy Awards for Best Foreign Language Film\newline The following 18 films, all from different countries, were
      & List of films submitted to the 80th Academy Awards for Best Foreign Language Film\newline The following 14 films, from 14 different countries,
      & 0.77 & 0.46 \\
    Vec2Text Base
      &
      & List of submissions to the 59th Academy Awards for Best Foreign Language Film. The following eight films, all from different countries, were
      & \textbf{0.91} & 0.76 \\
    Vec2Text + Corrector (50 Steps)
      &
      & Same as Vec2Text Base.
      & \textbf{0.91} & 0.76 \\
    \midrule
    \multicolumn{5}{l}{\textit{Out-of-distribution (OOD), Mistral-7B-v0.1 target}} \\
    \midrule
    \rowcolor{qblue} Ours
      & A 35-year-old female with abnormal heavy periods, fatigue, and dark spots on her hands and neck.
      & A 35-year-old woman with a BMI of 300 and a history of acne, hirsutism, and infertility.
      & \textbf{0.31} & 0.15 \\
    Vec2Text Base
      &
      & He was born in New York City and died in New York City in January 2013. During his stay at the YMCA,
      & 0.05 & 0.02 \\
    Vec2Text + Corrector (50 Steps)
      &
      & A rare condition characterized by a swollen thigh, a swollen thigh, a swollen thigh, and a swollen thigh, on
      & 0.11 & 0.04 \\
    \bottomrule
    \end{tabular}}
    % \vspace{-0.4cm}
    \end{table*}

\subsection{Inversion Performance Analysis}\label{sec:case}
To further evaluate Rep2Text and its usefulness for representation interpretation, we compare it with Vec2Text~\citep{morris2023text}, a strong embedding inversion baseline that uses a hypothesizer and a corrector model. 
Because our setting reconstructs text from last-token representations rather than sentence embeddings, we adapt Vec2Text to the same representation-inversion setting.

We train the Vec2Text hypothesizer and corrector on the same Wikipedia data used for Rep2Text, following the original setup and training each model for 10 epochs. 
Both methods invert last-token representations from Mistral-7B-v0.1, and Rep2Text uses Qwen-2.5-14B as the decoding model. 
We evaluate on held-out Wikipedia data and OOD clinical summaries containing patient names, ages, and symptoms, using an inversion length of 32 tokens. 
We use 32 tokens because the clinical summaries contain up to 25 tokens, so this length does not truncate clinical information.
Since the hypothesizer is the core inversion model in Vec2Text and the corrector mainly refines its output, we treat the hypothesizer as the primary baseline. 
Appendix~\ref{app:inver-perf} discusses the difference in decoder size between the two methods.
Additional details are provided in Appendix~\ref{app:ood}.

Table~\ref{tab:baseline} reports token-level, semantic, and LLM-judge metrics, including \textit{Token-F1} and \textit{BLEU} for comparison with Vec2Text. 
On in-distribution Wikipedia data, both methods achieve comparable semantic performance, while Rep2Text recovers more token-level information, improving \textit{Token F1} by 11\% and \textit{Entity Score} by 13\%. 
On OOD clinical data, Rep2Text generalizes substantially better, achieving 0.26 \textit{Token F1}, 0.37 \textit{ROUGE-1}, 0.64 \textit{Topic Score}, and 0.10 \textit{BLEU}, whereas Vec2Text Base reaches only 0.13 \textit{Token F1}, 0.14 \textit{ROUGE-1}, 0.21 \textit{Topic Score}, and 0.04 \textit{BLEU}.

To assess the reliability of the LLM-judge metrics, we further conduct a human evaluation check. 
As shown in Figure~\ref{fig:human_sanity}, GPT-4.1-mini produces scores in a similar range to human annotations, but tends to underestimate structure preservation and slightly overestimate entity and topic preservation. 
This suggests that LLM-as-a-judge is a useful scalable proxy, while its absolute scores should be interpreted with metric-specific caution. 
Detailed settings and discussion are provided in Appendix~\ref{app:human}.

We also evaluate Vec2Text with 50 corrector steps, in which the corrector iteratively refines the hypothesizer's output. 
The corrector provides only marginal gains on OOD data, despite converged training metrics. 
This may be because Vec2Text was designed for text encoder embeddings, while last-token representations may provide a less informative feedback signal for iterative editing. 
In contrast, Rep2Text directly aligns the representation space with the decoder embedding space, which appears more robust under distribution shift.

Table~\ref{tab:clinical} shows the outputs of the three methods on the same inputs. 
On in-distribution data, the methods show mixed results. 
Rep2Text is nearly exact on the first example, while Vec2Text performs better on the second. 
On OOD clinical data, Rep2Text better preserves the clinical context. 
In contrast, Vec2Text produces unrelated or repetitive text. 
Rep2Text can still hallucinate specific attributes, such as an implausible BMI. 
Thus, exact recovery of clinical facts remains imperfect. 
Overall, these examples agree with Table~\ref{tab:baseline}. 
The two methods perform similarly on in-distribution data, but Rep2Text generalizes better to OOD data.

\section{Conclusions}\label{sec:conclusion}
% In this work, we proposed Rep2Text, a novel framework for decoding full text from the last-token representation of LLMs, thereby quantifying what information these compressed representations preserve. Through comprehensive evaluations we revealed that different types of information are encoded at different network depths: structural information is best preserved in early-to-middle layers, while semantic information emerges progressively in middle-to-late layers. Our experiments demonstrate the feasibility of both same-model and cross-model representation inversion, with dimensional alignment playing a crucial role in cross-architecture decoding. We identified an information bottleneck effect where longer sequences lead to decreased inversion performance. Importantly, we showed that lightweight adapter training alone is sufficient to achieve strong inversion results, with the optimal configuration using projected token embeddings equal to the original sequence length. Looking forward, we plan to extend the Rep2Text framework to explain probing concept vectors and features learned by Sparse Autoencoders.
In this work, we explore the research question that to what extent can the original input text be recovered from a single last-token representation of LLMs? To answer this question, we proposed Rep2Text, a novel framework that employs a trainable adapter to project a target model's last-token representation into the embedding space of a decoding language model, which then autoregressively reconstructs the input text.
Our comprehensive evaluations indicate that roughly half of the tokens in 16-token sequences can be recovered from the compressed last-token representation while maintaining strong semantic integrity. Besides, experimental results show 
%an information bottleneck effect where 
longer sequences lead to decreased inversion performance, with reliable recovery achieved for sequences under 16 tokens. 
Additionally, Rep2Text also shows partial transfer to out-of-distribution clinical data, recovering coarse structural and semantic information, although exact recovery of critical attributes remains imperfect.
Rep2Text shows that a single intermediate representation can reveal part of the input text. Thus, deployed systems should protect intermediate representations. Our clinical evaluation uses LLM-generated summaries of a public dataset. Developing safeguards against representation inversion is future work.
%Looking forward, we plan to extend the Rep2Text framework to explain probing concept vectors and features learned by Sparse Autoencoders.
% \clearpage

% \section*{Limitations}
% Our Rep2Text framework has several limitations. First, our target models have at most 8B parameters. Our decoders scale up to 72B. In future work, we plan to invert representations from larger target LLMs. Second, due to the limited training resources available, we only use a sampled dataset from The Pile to fine-tune the inverter. Third, when applied to out-of-distribution medical notes, the inverter can suffer from domain misalignment, occasionally generating irrelevant medical website introductory text or failing to recover critical information. Fourth, GPT-4.1-mini underestimates structure preservation and slightly overestimates entity and topic preservation, as our human sanity check shows in Appendix~\ref{app:human}. Fifth, Rep2Text recovers about half of the tokens, and exact recovery of sensitive clinical attributes remains imperfect. Our clinical experiment measures generalization beyond the Wikipedia training data.

\bibliographystyle{unsrtnat}
\bibliography{ref}

\clearpage
\appendix
\counterwithin{figure}{section}
\counterwithin{table}{section}
% \onecolumn

% \includepdf[pages=-, landscape]{Token_level_embedding_reversion.pdf}
% Please add the following required packages to your document preamble:
% \usepackage{miultirow}

\section{Metrics}\label{app:metric}
All used metrics are defined as follows:

\paragraph{Token-level Accuracy.} We adopt ROUGE scores to measure the token-level accuracy~\citep{lin2004rouge}. Specifically, \textit{ROUGE-1}, and \textit{ROUGE-2} measure the recovery rate of individual tokens and 2-grams, respectively, while \textit{ROUGE-L} captures the longest common subsequence between the ground-truth and the inverted sequence. Detailed definitions of these metrics are as below:
\begin{itemize}
    \item \textbf{ROUGE-1 and ROUGE-2}: 
    For ROUGE-$k$, it computes the F-measure of $k$-grams extracted from a sequence. Suppose the set of $k$-grams is denoted as $G_k(S)$ and $G_k(\hat{k})$ for ground-truth sequence and inverted sequence respectively. ROUGE-$k$ is computed as follows:
\begin{equation}
\hspace{-0.1cm}\begin{array}{c}
\text{Overlap}_k=\displaystyle\sum_{g \in G_k(S) \cap G_k(\hat{S})} \min \left(\operatorname{cnt}_S(g), \operatorname{cnt}_{\hat{S}}(g)\right)\\
R_k=\frac{\text{Overlap}_k}{\left|G_k(S)\right|}, \quad P_k=\frac{\text{Overlap}_k}{\left|G_k(\hat{S})\right|}\\
\operatorname{ROUGE}_k=\frac{\left(1+\beta^2\right) R_k P_k}{R_k+\beta^2 P_k}
\end{array}
\end{equation}
where $\operatorname{cnt}(\cdot)$ denotes the count of the set.
\item \textbf{ROUGE-L}:
Give an ground-truth sequence $S=\langle s_1, \ldots, s_n\rangle$ and an inverted sequence $\hat{S}=\langle \hat{s}_1, \ldots, \hat{s}_m\rangle$, the length of their longest common subsequence is $LCS(S, \hat{S})$. The ROUGE-L score $F_{LCS}$ is defined as follows:
\begin{equation}\label{eq:rougel}
\begin{array}{c}
R_{L C S}=\frac{L C S(S, \hat{S})}{n}, \quad P_{L C S}=\frac{L C S(S, \hat{S})}{m}\\
F_{L C S}=\frac{\left(1+\beta^2\right) R_{L C S} P_{L C S}}{R_{L C S}+\beta^2 P_{L C S}}
\end{array}
\end{equation}
\end{itemize}

\paragraph{Sentence Structure and Entity Preservation.} To evaluate the preservation of sentence structure and entities, we use GPT-4.1-mini to rate the degree of preservation ranging on a 0-5 scale (normalized to 0-1), yielding the \textit{Structure Score} and \textit{Entity Score}, respectively. The structure score assesses how well the grammatical structure and sentence skeleton are preserved in the inverted sequences. While the entity score measures how accurately entity names and  their associated attributes are inverted. Detailed rating criteria are provided in Appendix~\ref{app:struct}.

\paragraph{Semantic Similarity.} We use BERTScore F1 and LLM-as-a-judge to collectively evaluate the semantic similarity between the ground-truth and inverted sequences~\citep{zhang2019bertscore}. BERTScore quantifies similarity in the embedding space, whereas the LLM-based evaluation measures topic relevance between ground-truth and inverted sequences. The scoring guidelines for LLM-as-a-judge evaluation are included in Appendix~\ref{app:struct}.

\section{Inverted Examples with Varying Token Lengths and Recovery Rate}\label{app:tokens}
As shown in Table~\ref{tab:token-examples}, for sequences with 8 tokens and 16 tokens, some inverted sequences fully recover the original text, while others fail to capture fine-grained details yet still preserve clear grammatical structures. For example, when inverting 16-token sequence, although the ROUGE-1 score is only 0.4, the original sentence ``\textit{Rob James may refer   to:\textbackslash{}n\textbackslash{}nRob James (singer) (}'' and the inverted sequence ``\textit{Mark Jones\textbackslash{}n\textbackslash{}nMark Jones may refer to:\textbackslash{}n\textbackslash{}nMark Jones (singer) (}'' share the same syntactic pattern, "[NAME]\textbackslash{}n\textbackslash{}n[NAME] (singer) (", and convey equivalent topic-level information. However, when the number of tokens exceed 16, the inverted sequences remain highly topic-relevant but tend to lose their global structural coherence, even when achieving a reasonable ROUGE-1 score.

% \begin{table}[!h]
%     \centering
%     \caption{Comparison of mean semantic scores between human and GPT-4.1-mini}
%     \begin{tabular}{c|c|c|c}\toprule
%          & Structure & Entity & Topic  \\\midrule
%          Human&$0.71_{\pm0.13}$& $0.49_{\pm 0.21}$& $0.67_{\pm 0.13}$ \\\midrule
%         GPT-4.1-mini &0.52& 0.57& 0.72 \\\bottomrule
%     \end{tabular}
%     \label{tab:human}
% \end{table}
% \setcounter{section}{1}

\section{Human Sanity Check}\label{app:human}
The reported human scores correspond to the mean ratings over the 200 sampled sequences. 
The five annotators are volunteers and receive no monetary compensation. 
To characterize the variability of human judgment, we also report the standard deviation across the mean scores of the five annotators. 
As shown in Figure~\ref{fig:human_sanity}, GPT-4.1-mini produces scores that are close to the range of human annotations across all three dimensions. 
In particular, its entity and topic scores are close to the human mean, while its structure score is lower than the human average but still comparable to the ratings assigned by several individual annotators. 
This suggests that GPT-4.1-mini does not behave as an outlier evaluator, but rather reflects one plausible annotation tendency within the range of human judgments.

At the same time, the comparison reveals metric-specific bias. 
GPT-4.1-mini is more conservative than human annotators when evaluating structure preservation, whereas it gives slightly higher scores for entity and topic preservation. 
Therefore, we use LLM-as-a-judge as a scalable proxy for semantic evaluation, while interpreting the absolute scores with caution, especially for structure preservation.

\section{More Results on Inversion Performance Analysis}\label{app:inver-perf}
We compare Rep2Text with the adapted Vec2Text baseline under the last-token representation inversion setting. 
For Vec2Text, both the hypothesizer and corrector follow the original two-stage design with T5-base, using a learning rate of $10^{-3}$ and a warmup ratio of 0.10. 
For Rep2Text, we use Qwen-2.5-14B as the decoding model and train only the adapter while keeping the decoder frozen, using a learning rate of $1.5\times10^{-3}$ and a warmup ratio of 0.30. 
All runs are conducted on two A100 GPUs.

The two methods use decoders of different sizes.
Vec2Text uses T5-base, while Rep2Text uses Qwen-2.5-14B.
However, decoder size has a limited effect on Rep2Text.
In the 16-token setting, the Llama-3.2-3B decoder inverts Mistral-7B-v0.1 representations at least as well as Qwen-2.5-14B on all metrics, as shown in Table~\ref{tab:model}.
Meanwhile, Rep2Text outperforms Vec2Text Base by a large margin on OOD clinical data.
It reaches 0.37 ROUGE-1 and 0.64 topic score, while Vec2Text Base reaches 0.14 and 0.21.
Thus, the gap is unlikely to come from decoder capacity alone.

\begin{figure*}[!t]
\centering
\subfloat[]{
    \includegraphics[width=0.32\linewidth]{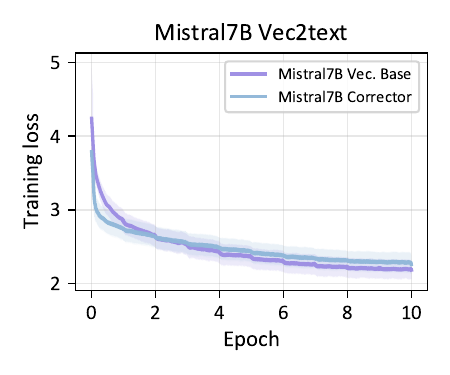}
}
\hfill
\subfloat[]{
    \includegraphics[width=0.32\linewidth]{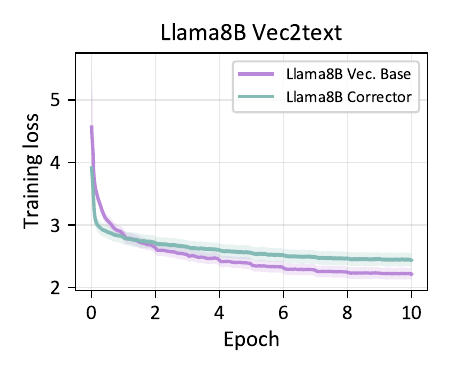}
}
\hfill
\subfloat[]{
    \includegraphics[width=0.32\linewidth]{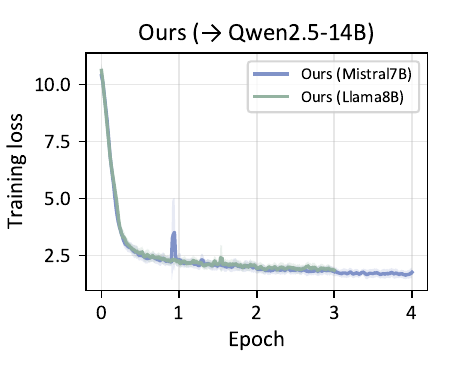}
}
\caption{Training dynamics of both Vec2text and our methods.}
\label{fig:training_dynamic}
\end{figure*}

Figure~\ref{fig:training_dynamic} shows that the Vec2Text corrector reduces training loss, but this does not translate into better inversion performance. 
As shown in Table~\ref{tab:baseline}, adding the corrector brings no consistent improvement and even degrades the base model, especially in the OOD clinical setting. 
This suggests that the corrector does not learn an effective refinement strategy for last-token representation inversion.

In contrast, Rep2Text rapidly reduces training loss and achieves stronger downstream inversion performance. 
This supports our hypothesis that directly aligning the target representation space with the decoder embedding space is more effective than iteratively editing decoded hypotheses when the input signal is a compressed last-token representation.

\section{Results on Representation-Text Correspondence Analysis}\label{app:random}
We conduct three experiments to examine whether Rep2Text relies on the correspondence between representations and reconstruction targets. 
All settings use Mistral-7B-v0.1 as the target and Llama-3.2-3B as the decoder. 
The maximum sequence length is set to 16 tokens, and the number of projected vectors is set to 16. 
Experiments are trained with an effective batch size of 512, a learning rate of $7\times10^{-4}$, a warmup ratio of 0.15, and 4 epochs on two A100 GPUs.

In the \textit{Baseline} setting, each representation is paired with its original input sequence. 
In the \textit{Reverse} setting, we reverse the input sequence at the word level and then retokenize it as the reconstruction target. 
In the \textit{Random} setting, each representation is paired with an unrelated target sequence, thereby breaking the representation--text correspondence.

\begin{figure*}[t]
    \centering
    \subfloat[Loss]{
        \includegraphics[width=0.23\linewidth]{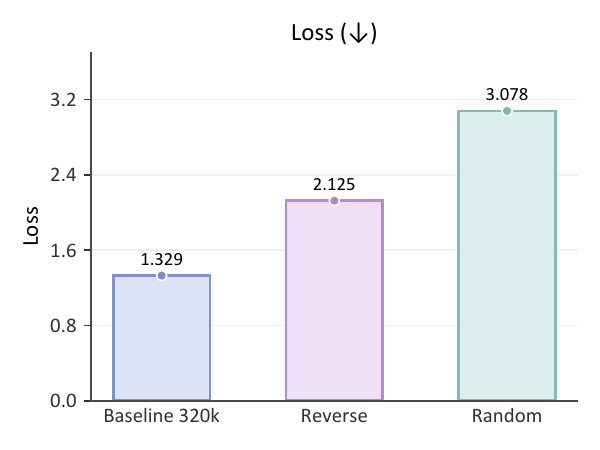}
    }
    \hfill
    \subfloat[ROUGE-1]{
        \includegraphics[width=0.23\linewidth]{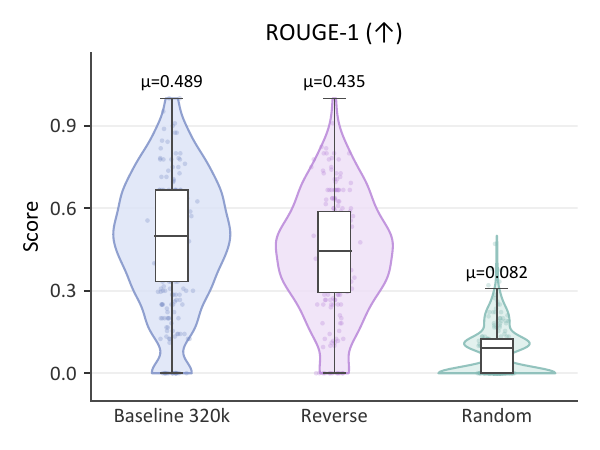}
    }
    \hfill
    \subfloat[ROUGE-2]{
        \includegraphics[width=0.23\linewidth]{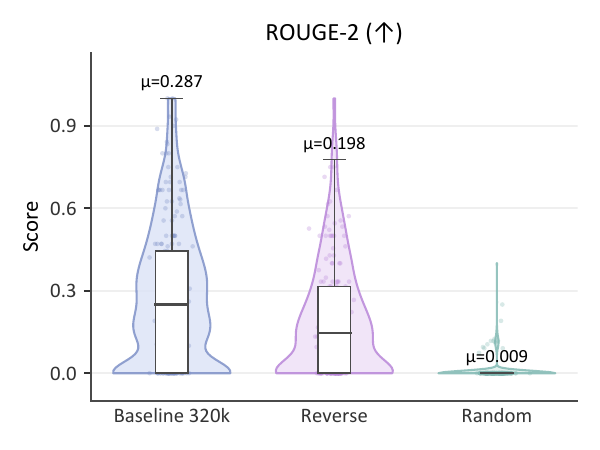}}
    \hfill
    \subfloat[ROUGE-L]{
        \includegraphics[width=0.23\linewidth]{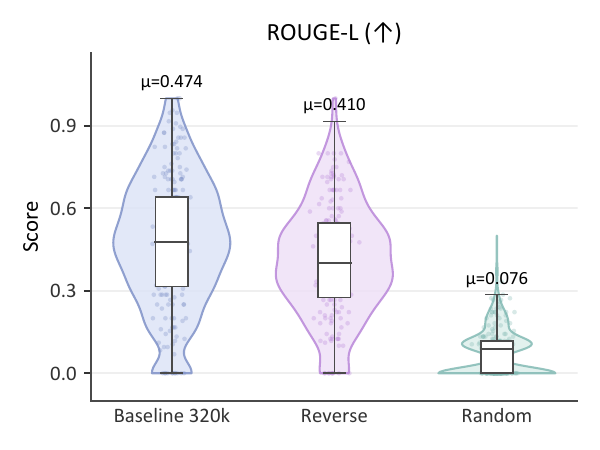}
    }\\
    \subfloat[Perplexity]{
        \includegraphics[width=0.23\linewidth]{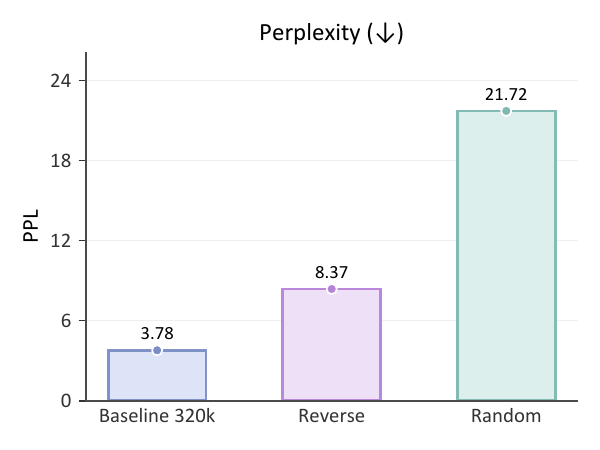}}
    \hfill        
    \subfloat[Token F1]{
        \includegraphics[width=0.23\linewidth]{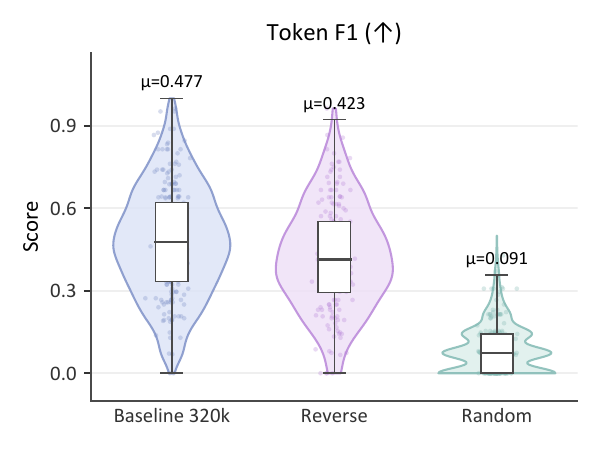}}
    \hfill
    \subfloat[BLEU]{
        \includegraphics[width=0.24\linewidth]{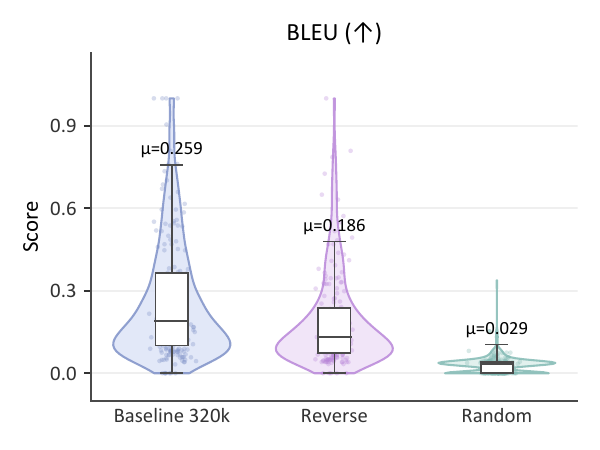}}
    \hfill
    \subfloat[BERTScore F1]{
        \includegraphics[width=0.24\linewidth]{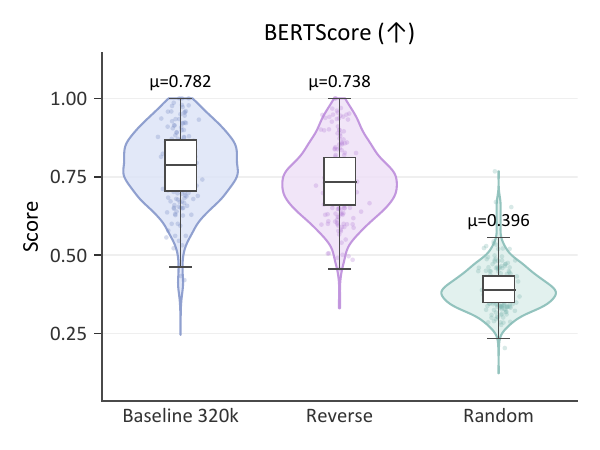}}
    \caption{Performance comparison under different representation--text correspondence settings. 
The baseline uses correctly paired representations and original input sequences, Reverse uses word-level reversed input sequences as reconstruction targets, and Random randomly pairs representations with unrelated target sequences. 
The strong degradation under Random indicates that Rep2Text relies on representation-specific information rather than only decoder language priors.}\label{fig:random}
\end{figure*}

As shown in Figure~\ref{fig:random}, \textit{Random} performs substantially worse than \textit{Baseline}: the training loss increases from 1.329 to 3.078, while ROUGE-L, Token F1, and BERTScore decrease from 0.474, 0.477, and 0.782 to 0.076, 0.091, and 0.396, respectively. 
This confirms that the inverter cannot learn meaningful reconstruction when representations are paired with unrelated text.

By contrast, \textit{Reverse} achieves 0.410 ROUGE-L, 0.423 Token F1, and 0.738 BERTScore. 
Although it underperforms the original-order \textit{Baseline}, it remains far above \textit{Random}, suggesting that the model can recover input-specific information even when the target follows an unnatural word-order.

\section{Layer-wise Invertibility across Target Models}\label{app:layer}
We sweep the layers of Mistral-7B-v0.1, Gemma-7B, and Llama-3.2-3B at roughly matched relative depths.
All runs use Llama-3.1-8B as the decoder and 16-token Wikipedia sequences.
Table~\ref{tab:layer} reports the results together with the Llama-3.1-8B sweep from $\S$\ref{sec:layerwise}.
Mistral-7B-v0.1 performs best around its middle layers.
Llama-3.2-3B performs similarly from layer 4 to layer 13 and degrades in later layers.
Gemma-7B performs best in its late layers.
Thus, the best inversion layer depends on the model architecture.

\begin{table}[h]
\centering
\caption{Layer-wise inversion performance of four target models with Llama-3.1-8B as the decoder. Bold rows mark the layer with the highest ROUGE-1, which Table~\ref{tab:model} uses. The L10 rows of Mistral-7B-v0.1, Gemma-7B, and Llama-3.2-3B come from the main runs in Table~\ref{tab:model}.}
\label{tab:layer}
\resizebox{\linewidth}{!}{
\begin{tabular}{lccccccc}
\toprule
\textbf{Layer} & \textbf{ROUGE-1} & \textbf{ROUGE-2} & \textbf{ROUGE-L} & \textbf{BERTScore} & \textbf{Structure} & \textbf{Entity} & \textbf{Topic} \\
\midrule
\multicolumn{8}{l}{\textit{Llama-3.1-8B (32 layers)}} \\
L5 & 0.46 & 0.26 & 0.44 & 0.76 & 0.62 & 0.69 & 0.88 \\
\textbf{L10} & \textbf{0.48} & \textbf{0.28} & \textbf{0.47} & \textbf{0.78} & \textbf{0.66} & \textbf{0.74} & \textbf{0.91} \\
L15 & 0.48 & 0.28 & 0.45 & 0.77 & 0.62 & 0.75 & 0.91 \\
L20 & 0.44 & 0.24 & 0.42 & 0.74 & 0.56 & 0.72 & 0.91 \\
L25 & 0.42 & 0.21 & 0.39 & 0.72 & 0.54 & 0.71 & 0.89 \\
L30 & 0.40 & 0.21 & 0.37 & 0.71 & 0.53 & 0.69 & 0.88 \\
\midrule
\multicolumn{8}{l}{\textit{Mistral-7B-v0.1 (32 layers)}} \\
L5 & 0.52 & 0.32 & 0.50 & 0.79 & 0.67 & 0.71 & 0.89 \\
L10 & 0.52 & 0.32 & 0.51 & 0.81 & 0.71 & 0.75 & 0.90 \\
\textbf{L15} & \textbf{0.55} & \textbf{0.35} & \textbf{0.53} & \textbf{0.81} & \textbf{0.70} & \textbf{0.75} & \textbf{0.92} \\
L20 & 0.53 & 0.33 & 0.52 & 0.79 & 0.64 & 0.74 & 0.92 \\
L25 & 0.49 & 0.29 & 0.47 & 0.76 & 0.60 & 0.72 & 0.90 \\
L30 & 0.47 & 0.26 & 0.45 & 0.75 & 0.60 & 0.72 & 0.91 \\
\midrule
\multicolumn{8}{l}{\textit{Gemma-7B (28 layers)}} \\
L4 & 0.45 & 0.24 & 0.43 & 0.72 & 0.63 & 0.56 & 0.76 \\
L9 & 0.53 & 0.31 & 0.51 & 0.76 & 0.68 & 0.59 & 0.80 \\
L10 & 0.51 & 0.28 & 0.49 & 0.75 & 0.66 & 0.59 & 0.79 \\
L13 & 0.53 & 0.31 & 0.51 & 0.76 & 0.68 & 0.58 & 0.79 \\
L18 & 0.50 & 0.28 & 0.49 & 0.74 & 0.68 & 0.56 & 0.77 \\
L22 & 0.57 & 0.35 & 0.55 & 0.77 & 0.68 & 0.61 & 0.82 \\
\textbf{L26} & \textbf{0.64} & \textbf{0.42} & \textbf{0.61} & \textbf{0.80} & \textbf{0.71} & \textbf{0.64} & \textbf{0.86} \\
\midrule
\multicolumn{8}{l}{\textit{Llama-3.2-3B (28 layers)}} \\
L4 & 0.45 & 0.25 & 0.43 & 0.75 & 0.60 & 0.67 & 0.87 \\
\textbf{L9} & \textbf{0.46} & \textbf{0.25} & \textbf{0.44} & \textbf{0.76} & \textbf{0.64} & \textbf{0.72} & \textbf{0.89} \\
L10 & 0.45 & 0.25 & 0.43 & 0.76 & 0.64 & 0.72 & 0.88 \\
L13 & 0.46 & 0.25 & 0.44 & 0.76 & 0.62 & 0.73 & 0.90 \\
L18 & 0.42 & 0.22 & 0.39 & 0.72 & 0.55 & 0.71 & 0.90 \\
L22 & 0.39 & 0.19 & 0.37 & 0.71 & 0.53 & 0.70 & 0.89 \\
L26 & 0.38 & 0.18 & 0.36 & 0.70 & 0.53 & 0.69 & 0.88 \\
\bottomrule
\end{tabular}}
\end{table}

\section{Inversion of Longer Sequences}\label{app:long}
We extend the sequence-length analysis in $\S$\ref{sec:token-length} to 128 and 256 tokens.
These runs use Llama-3.2-3B as both the target and decoder.
We also train a 16-token model under the same setting as an anchor.
Table~\ref{tab:long} reports the results.

Lexical and structural recovery drop sharply as inputs grow.
Under the 3B setting, ROUGE-1 drops from 0.46 at 16 tokens to 0.30 at 128 tokens and stays at 0.30 at 256 tokens.
ROUGE-2, ROUGE-L, entity, and structure scores keep declining.
Semantic metrics drop much more slowly.
At 256 tokens, the topic score is 0.68 and BERTScore is 0.63.
This BERTScore is well above the 0.40 of the Random control in Appendix~\ref{app:random}.
Thus, a single last-token representation loses most of the exact wording of long inputs but keeps their coarse topic.

\begin{table}[h]
\centering
\caption{Inversion performance across input lengths. 8B denotes Llama-3.1-8B and 3B denotes Llama-3.2-3B. Each model serves as both the target and the decoder.}
\label{tab:long}
\resizebox{\linewidth}{!}{
\begin{tabular}{llccccccc}
\toprule
\textbf{Tokens} & \textbf{Model} & \textbf{ROUGE-1} & \textbf{ROUGE-2} & \textbf{ROUGE-L} & \textbf{BERTScore} & \textbf{Structure} & \textbf{Entity} & \textbf{Topic} \\
\midrule
8 & 8B & 0.62 & 0.42 & 0.59 & 0.83 & 0.78 & 0.77 & 0.93 \\
16 & 8B & 0.48 & 0.28 & 0.47 & 0.78 & 0.66 & 0.74 & 0.91 \\
32 & 8B & 0.40 & 0.18 & 0.36 & 0.72 & 0.57 & 0.68 & 0.86 \\
64 & 8B & 0.36 & 0.13 & 0.28 & 0.68 & 0.49 & 0.60 & 0.82 \\
\midrule
16 & 3B & 0.46 & 0.26 & 0.45 & 0.76 & 0.64 & 0.69 & 0.88 \\
128 & 3B & 0.30 & 0.08 & 0.22 & 0.63 & 0.37 & 0.41 & 0.72 \\
256 & 3B & 0.30 & 0.07 & 0.20 & 0.63 & 0.36 & 0.39 & 0.68 \\
\bottomrule
\end{tabular}}
\end{table}

\section{More Details on OOD Experiments}\label{app:ood}
\subsection{OOD Dataset}\label{app:ood-data}
We use an open-source clinical dataset from Hugging Face\footnote{https://huggingface.co/datasets/Sadaftb/clinical-nlp-patient-notes}. 
To fit patient histories within 32 tokens, we use Qwen-2.5-32B to summarize 2,500 randomly sampled patient notes. 
The instruction is:
\begin{tcolorbox}[
    enhanced,
    breakable,
    colback=orange!10!white, colframe=blue!5!black,
    arc=2mm,
    boxrule=1pt,
    title=,
    coltitle=white,
    attach boxed title to top left={yshift=-2mm, xshift=3mm},
    top=0.5mm,
    left=1mm,
    right=1mm,
    bottom=0.5mm,
]
\setlength{\parskip}{2pt}
You're given a piece of clinical note in the following context. Summarize the clinical note into ONE fluent sentence (<=25 tokens). Must include: patient name, date of birth, gender, and key symptoms. Start with "[ans]". Output only the summary.

\{note\}
\end{tcolorbox}

\subsection{Experimental Details on Vec2Text}
We adapted the original Vec2Text into inverting last-token representation. Here, we use Mistral-7B-v0.1 as our embedding model. The training of both base model and corrector adopts default training parameters of Vec2Text. To make sure the training is converged well without overfitting, we set up both training process to be 10 epochs which takes more than 18 hours separately. Base model is set to be t5-base, num repeat tokens are 16 and max sequence length is set to be 64. 

\subsection{Inverted Examples with OOD Clinical Notes}
Table~\ref{tab:clinical} compares the three methods on the same inputs.

\section{More Implementation Details}
\label{app:training}
This section provides additional implementation details and experiment-specific hyperparameters omitted from the main text. Unless otherwise specified, all experiments use the same default training configuration described below.

Unless otherwise specified, all experiments are trained on Wikipedia with 640K training examples, using embeddings extracted from layer 10. We train for 3 epochs with a cosine learning rate scheduler and a warmup ratio of 0.15. The default learning rate is $10^{-3}$, except where noted otherwise. We use Adam with $\beta=(0.9, 0.95)$, weight decay $0.01$, label smoothing $0.075$, and adapter dropout $0.1$. Unless otherwise stated, the number of projected token vectors is set equal to the number of tokens being inverted, as this gives the best performance (Appendix~\ref{app:ablation}), and the hidden expansion factor is fixed at $0.5$. Most experiments are run on two NVIDIA A100 GPUs (80GB each), with a global batch size of 1024 implemented through per-GPU batch size and gradient accumulation. More detailed setups are shown in Table~\ref{tab:exp_settings}.

\begin{table*}[!h]
\centering
\caption{Experiment-specific hyperparameters for transfer, token-length, and layer-sweep experiments. All unspecified settings follow the shared default configuration described in Appendix~\ref{app:training}. The Qwen2.5-72B run uses a tensor-parallel size of 4, so its 8 GPUs form 2 data-parallel replicas.}
\label{tab:exp_settings}
\setlength{\tabcolsep}{3pt}
\scalebox{0.8}{
\begin{tabular}{ l l c c c c c c cc}
\toprule
\textbf{Emb Model} & \textbf{Dec Model} & \textbf{Max Len} & $\mathbf{n}_{\textbf{vecs}}$ & \textbf{\#GPU} & \textbf{Mini batch} & \textbf{Grad Accum} & \textbf{Eff. Batch} & \textbf{LR} & \textbf{Warmup Ratio}\\
\midrule
\multicolumn{10}{l}{\textit{Transfer experiments}} \\
\midrule
Gemma-7B     & Llama-3.1-8B  & 16 & 16 & 2 & 32 & 16 & 1024 & $1\mathrm{e}{-3}$ & 0.15\\
Llama-3.2-3B & Llama-3.1-8B  & 16 & 16 & 2 & 32 & 16 & 1024 & $1\mathrm{e}{-3}$ & 0.15\\
Mistral-7B   & Llama-3.1-8B  & 16 & 16 & 2 & 32 & 16 & 1024 & $1\mathrm{e}{-3}$ & 0.15\\
Mistral-7B   & Llama-3.2-3B  & 16 & 16 & 2 & 64 & 8  & 1024 & $1\mathrm{e}{-3}$ & 0.15\\
Mistral-7B   & Qwen2.5-14B   & 16 & 16 & 2 & 32 & 16 & 1024 & $1\mathrm{e}{-3}$ & 0.30\\
Mistral-7B & Qwen2.5-14B & 32 & 32 & 4 & 16 & 8 & 512 & $1.5\mathrm{e}{-3}$ & 0.30\\
Mistral-7B & Qwen2.5-32B & 16 & 16 & 8 & 28 & 9 & 2016 & $1.5\mathrm{e}{-3}$ & 0.30\\
Mistral-7B & Qwen2.5-72B & 16 & 16 & 8 & 64 & 8 & 1024 & $1\mathrm{e}{-3}$ & 0.15\\
\midrule
\multicolumn{10}{l}{\textit{Token-length experiments (Llama-3.1-8B $\leftrightarrow$ Llama-3.1-8B)}} \\
\midrule
Llama-3.1-8B & Llama-3.1-8B & 8   & 8   & 2 & 32 & 16 & 1024 & $1\mathrm{e}{-3}$ & 0.15\\
Llama-3.1-8B & Llama-3.1-8B & 16  & 16  & 2 & 32 & 16 & 1024 & $1\mathrm{e}{-3}$ & 0.15\\
Llama-3.1-8B & Llama-3.1-8B & 32  & 32  & 2 & 32 & 16 & 1024 & $1\mathrm{e}{-3}$ & 0.15\\
Llama-3.1-8B & Llama-3.1-8B & 64  & 64  & 4 & 16 & 16 & 1024 & $3\mathrm{e}{-3}$ & 0.15\\
\midrule
\multicolumn{10}{l}{\textit{Token-length experiments (Llama-3.2-3B $\leftrightarrow$ Llama-3.2-3B)}} \\
\midrule
Llama-3.2-3B & Llama-3.2-3B & 128 & 128 & 2 & 16 & 32 & 1024 & $1\mathrm{e}{-3}$ & 0.15\\
Llama-3.2-3B & Llama-3.2-3B & 256 & 256 & 2 & 8  & 64 & 1024 & $1\mathrm{e}{-3}$ & 0.15\\
\midrule
\multicolumn{10}{l}{\textit{Layer sweeps (all swept layers)}} \\
\midrule
Mistral-7B   & Llama-3.1-8B  & 16 & 16 & 2 & 32 & 16 & 1024 & $1\mathrm{e}{-3}$ & 0.15\\
Gemma-7B     & Llama-3.1-8B  & 16 & 16 & 2 & 32 & 16 & 1024 & $1\mathrm{e}{-3}$ & 0.15\\
Llama-3.2-3B & Llama-3.1-8B  & 16 & 16 & 2 & 32 & 16 & 1024 & $1\mathrm{e}{-3}$ & 0.15\\
\bottomrule
\end{tabular}}
\end{table*}

\section{Ablation Study on Adapter Hidden Dimensions}\label{app:ablation}

We study how adapter design affects inversion performance by varying the hidden and output dimensions of the two-layer MLP adapter. All experiments are conducted on 16-token sequences using layer-10 representations from Llama-3.1-8B.

\begin{figure*}[!t]
\centering
\subfloat[]{
    \includegraphics[width=0.47\linewidth]{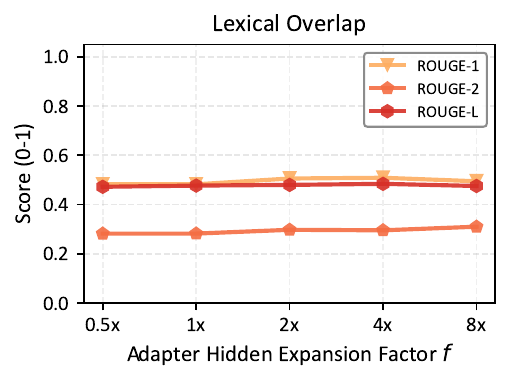}
    \label{fig:exp-lexcial}
}
\hfill
\subfloat[]{
    \includegraphics[width=0.47\linewidth]{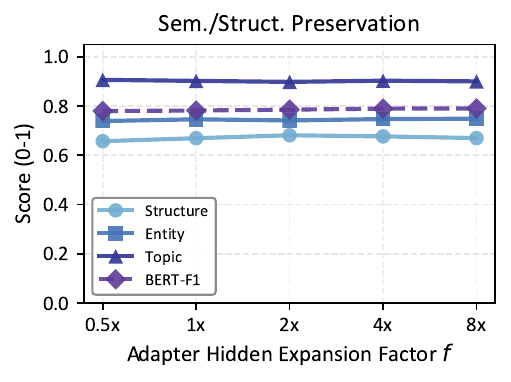}
    \label{fig:exp-semantic}
}
\caption{Inversion Performance on varying expansion factors.}
\label{fig:hidden}
\end{figure*}

As defined in Section~\ref{sec:inver-design}, the hidden dimension of the first adapter layer is $d^{\mathrm{hid}} = f \cdot d$, where $d$ is the input representation dimension and $f$ is an expansion factor. To evaluate the effect of hidden-layer capacity, we vary $f \in \{0.5, 1, 2, 4, 8\}$ while keeping the second-layer output dimension fixed. As shown in Figure~\ref{fig:hidden}, inversion performance is largely robust to changes in the hidden dimension. Semantic-level recovery is nearly unchanged, while token-level recovery, measured by ROUGE-1, improves slightly as the hidden dimension increases. This suggests that a larger hidden layer may provide additional capacity for recovering lexical information, but is not critical for preserving semantic content.

\begin{figure*}[!t]
\centering
\subfloat[]{
    \includegraphics[width=0.47\linewidth]{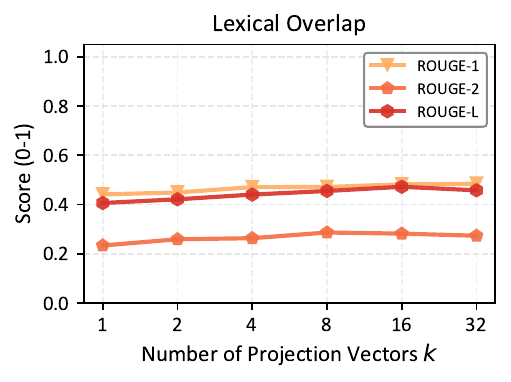}
    \label{fig:nvec-lexcial}
}
\hfill
\subfloat[]{
    \includegraphics[width=0.47\linewidth]{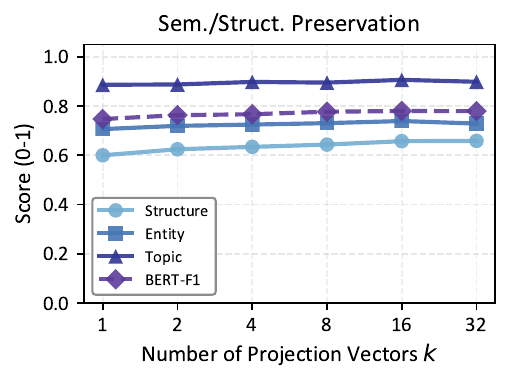}
    \label{fig:nvec-semantic}
}
\caption{Inversion performance on different number of inverted embedding tokens.}
\label{fig:nvec}
\end{figure*}

We further vary the output dimension of the second adapter layer. Specifically, we set the output dimension to $k \cdot d'$, where $d'$ is the embedding dimension of the decoder. This corresponds to projecting the target representation into $k$ token-embedding slots. We vary $k \in \{1, 2, 4, 8, 16, 32\}$ to test whether more projected embeddings help recover more information from the last-token representation. As shown in Figure~\ref{fig:nvec}, both token-level and semantic-level performance improve as $k$ increases up to $16$. However, performance degrades when $k=32$, which exceeds the ground-truth sequence length. This suggests that adding more projected embeddings than the target length may introduce redundant or noisy information.

Overall, the first-layer expansion factor has limited impact on inversion performance, whereas the number of projected token embeddings is more important. Performance is strongest when the number of projected embeddings matches the ground-truth token length. Based on these findings, we use $f=0.5$ and set the number of projected embeddings to the ground-truth token length in our main experiments.

\section{Training Dataset Size Ablation}~\label{app:dataset}
To examine whether Rep2Text benefits smoothly from additional training data, we train adapters with 10K, 50K, 100K, 320K, and 640K Wikipedia examples while adjusting the effective batch size and learning rate using square-root scaling. The 640K setting corresponds to the default configuration used in the main experiments.
\begin{figure*}[h]
    \centering
    \subfloat[Loss]{
        \includegraphics[width=0.4\linewidth]{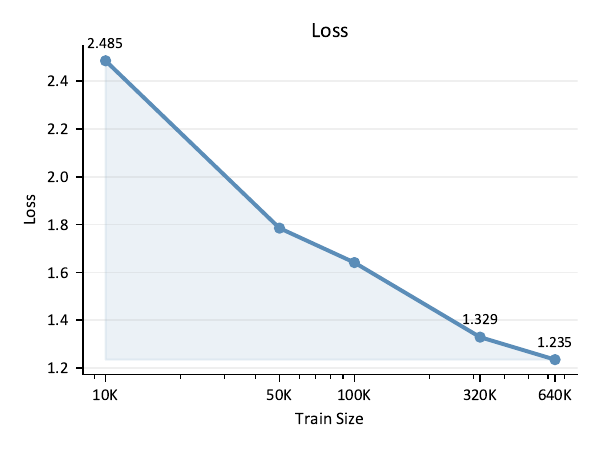}\label{fig:dataset-size-loss}}
    \hfill
    \subfloat[Perplexity]{
        \includegraphics[width=0.4\linewidth]{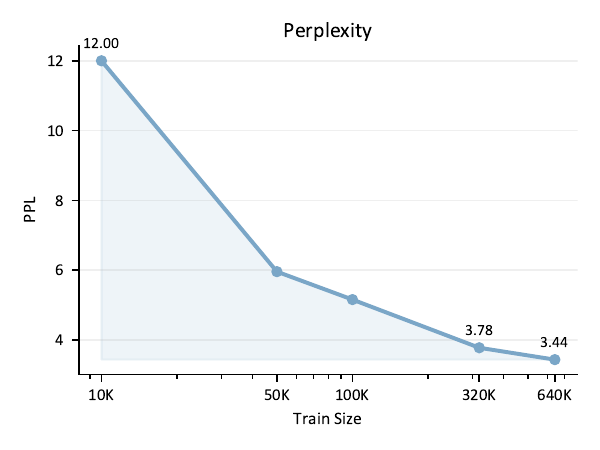}
        \label{fig:dataset-size-ppl}
    }
    \hfill
    \subfloat[BLEU]{
        \includegraphics[width=0.4\linewidth]{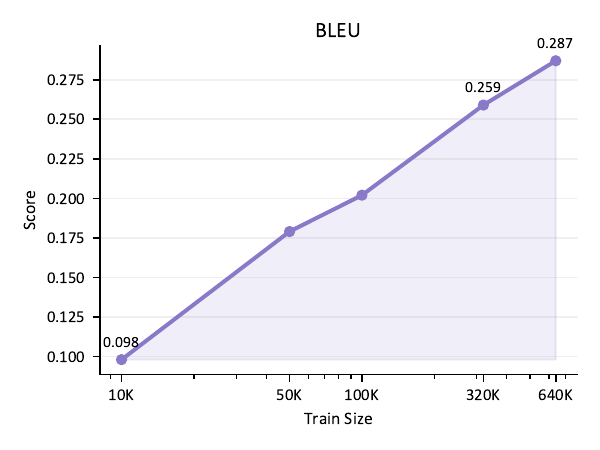}
        \label{fig:dataset-size-bleu}
    }\hfill
    \subfloat[BERTScore]{
        \includegraphics[width=0.4\linewidth]{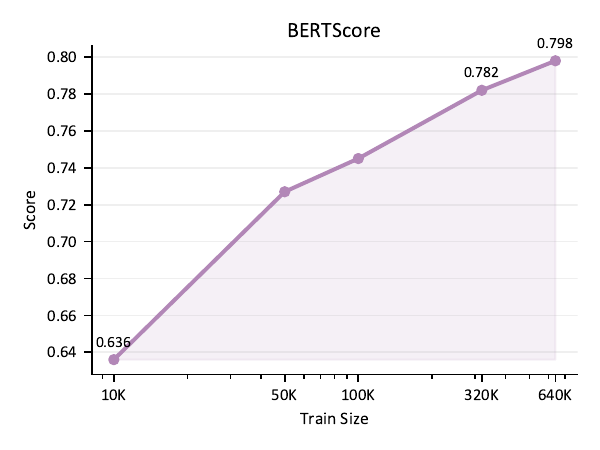}
        \label{fig:dataset-size-bertscore}
    }
    \\
    \subfloat[Token F1]{
        \includegraphics[width=0.4\linewidth]{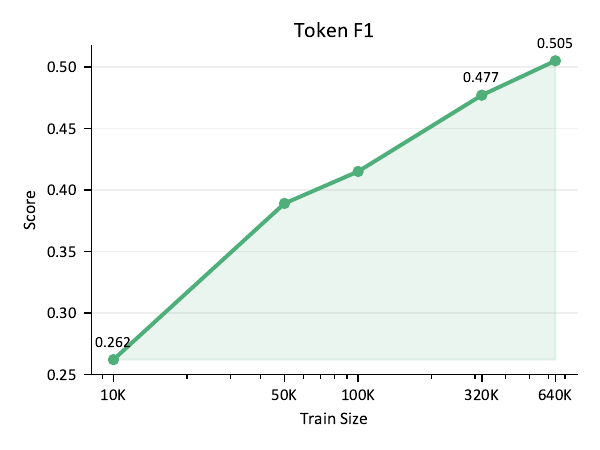}
        \label{fig:dataset-size-token-f1}
    }
    \hfill
    \subfloat[ROUGE-1]{
        \includegraphics[width=0.4\linewidth]{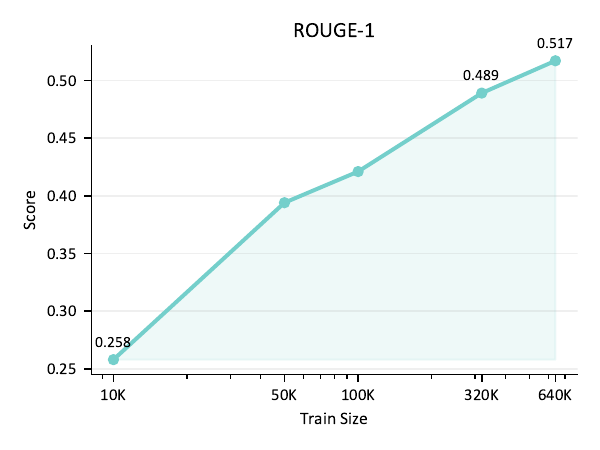}
        \label{fig:dataset-size-rouge-1}
    }
    \hfill
    \subfloat[ROUGE-2]{
        \includegraphics[width=0.4\linewidth]{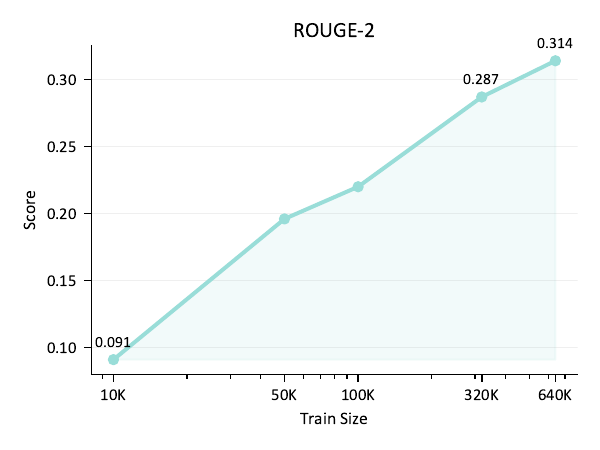}
        \label{fig:dataset-size-rouge-2}
    }
    \hfill
    \subfloat[ROUGE-L]{
        \includegraphics[width=0.4\linewidth]{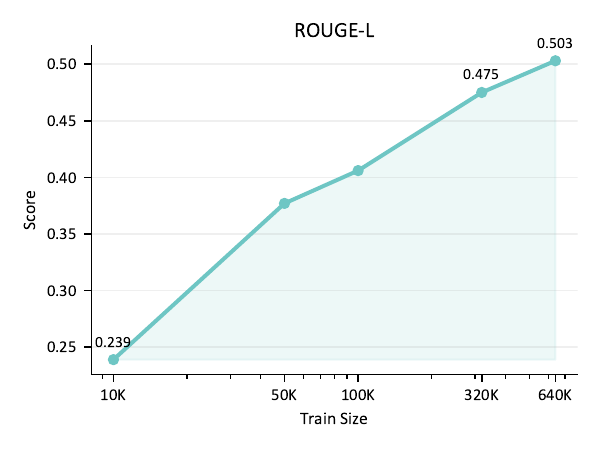}
        \label{fig:dataset-size-rouge-l}
    }
    \caption{Dataset size ablation across eight metrics. Increasing the number of training examples consistently improves Rep2Text inversion performance: loss and perplexity decrease, while token-level and semantic reconstruction metrics increase. The gains remain positive from 320K to 640K examples, but become smaller, indicating diminishing returns at larger data scales.}
    \label{fig:dataset-size-ablation}
\end{figure*}

\begin{table}[h]
\centering
\caption{Training configurations for the dataset size ablation.}
\label{tab:dataset_size_setup}
\begin{tabular}{lcccccc}
\toprule
Train Size & Epochs & Per-GPU BS & Grad. Accum. & GPUs & Eff. BS & LR \\
\midrule
10K  & 15 & 32 & 2 & 1 & 64   & $3\times 10^{-4}$ \\
50K  & 6  & 32 & 4 & 1 & 128  & $4\times 10^{-4}$ \\
100K & 5  & 32 & 4 & 2 & 256  & $5\times 10^{-4}$ \\
320K & 4  & 64 & 4 & 2 & 512  & $7\times 10^{-4}$ \\
640K & 3  & 64 & 8 & 2 & 1024 & $1\times 10^{-3}$ \\
\bottomrule
\end{tabular}
\end{table}

The experimental setups are shown in Table~\ref{tab:dataset_size_setup}. The learning rate is scaled according to $\sqrt{\text{effective batch size}/1024}\times 10^{-3}$. Unless otherwise specified, all models are trained under the same setting as the main experiments: we use 16-token Wikipedia sequences, extract layer-10 last-token representations, and fine-tune only the adapter while keeping the decoder frozen.

The results in Figure~\ref{fig:dataset-size-ablation} show a clear monotonic trend across all metrics: as the number of training examples increases, loss and perplexity decrease, while token-level and semantic reconstruction metrics improve consistently. In particular, BERTScore increases from 0.636 at 10K examples to 0.798 at 640K examples, suggesting that larger training sets help the adapter learn a more effective mapping from last-token representations to the decoder embedding space. At the same time, the marginal gains shrink at larger data scales. For example, increasing the dataset from 320K to 640K improves BERTScore only from 0.782 to 0.798 and Token F1 from 0.477 to 0.505, which is smaller than the improvements observed in earlier scaling regimes. This indicates that the data scaling curve starts to flatten at the 640K examples. Therefore, we adopt 640K as the default setting in the main experiments, as it offers a favorable balance between inversion quality and training cost.

\section{Instruction Examples}\label{app:prompt}
We include prompts for GPT-4.1-mini as below.

\subsection{Scoring Instructions}\label{app:struct}
\begin{tcolorbox}[
    enhanced,
    breakable,
    colback=orange!10!white, colframe=blue!5!black,
    arc=2mm,
    boxrule=1pt,
    title={\bfseries Structure Score},
    coltitle=white,
    attach boxed title to top left={yshift=-2mm, xshift=3mm},
    boxed title style={enhanced, colback=blue!5!black, colframe=blue!5!black, arc=2mm, boxrule=0pt},
    top=0.5mm, left=1mm, right=1mm, bottom=0.5mm,
]
\setlength{\parskip}{4pt}
\small
\vskip8pt
\textbf{You are given two sentences:}
\begin{itemize}[nosep, topsep=2pt, itemsep=0pt, parsep=0pt,leftmargin=1.2cm]
    \item[\textbf{[GT]:}] \texttt{self.gt\_sen}
    \item[\textbf{[GEN]:}] \texttt{self.gen\_sen}
\end{itemize}
Evaluate \textbf{Structural Frame Similarity}. Focus on whether [GEN] preserves the core grammatical structure and sentence skeleton of [GT]: same basic clause structure, same number/type of major phrases, similar syntactic relationships, consistent verb tense/aspect.

% \begin{tcolorbox}[enhanced, colback=gray!5, colframe=blue!5!black, arc=2mm, boxrule=1pt, left=1mm, right=1mm, top=2mm, bottom=2mm]
\textbf{Scoring:}
\begin{itemize}[nosep, topsep=2pt, itemsep=0pt, parsep=0pt, style=nextline, leftmargin=0.7cm]
    \item[\textbf{5}:] Identical structure — same words in same order
    \item[\textbf{4}:] Nearly identical — same pattern with entity substitutions or minor reordering
    \item[\textbf{3}:] Moderately similar — core structure maintained but notable changes (e.g., active to passive)
    \item[\textbf{2}:] Somewhat similar — recognizable elements but significant differences
    \item[\textbf{1}:] Minimally similar — only basic sentence type matches
    \item[\textbf{0}:] Completely different structures
\end{itemize}
% \end{tcolorbox}

\textbf{Answer:} \texttt{[ANS] structure: [score]/5}
\end{tcolorbox}

\begin{tcolorbox}[
    enhanced,
    breakable,
    colback=orange!10!white, colframe=blue!5!black,
    arc=2mm,
    boxrule=1pt,
    title={\bfseries Entity Score (Clinical)},
    coltitle=white,
    attach boxed title to top left={yshift=-2mm, xshift=3mm},
    boxed title style={enhanced, colback=blue!5!black, colframe=blue!5!black, arc=2mm, boxrule=0pt},
    top=0.5mm, left=1mm, right=1mm, bottom=0.5mm,
]
\setlength{\parskip}{4pt}
\small\vskip8pt
\textbf{You are given two sentences:}
\begin{itemize}[nosep, topsep=2pt, itemsep=0pt, parsep=0pt, leftmargin=1.2cm]
    \item[\textbf{[GT]:}] \texttt{self.gt\_sen}
    \item[\textbf{[GEN]:}] \texttt{self.gen\_sen}
\end{itemize}
Evaluate the \textbf{Entity/Role Consistency and Plausibility} between these snippets.

\textbf{Clinical Entity Type Matching Rules:}
\begin{itemize}[nosep, topsep=2pt, itemsep=0pt, parsep=0pt, style=nextline, leftmargin=0.7cm]
    \item[-] {Patient Demographics}: Age (±5 years = high), same biological sex
    \item[-] {Symptoms / Chief Complaint}: Same symptom category; penalise crossing body systems
    \item[-] {Diagnosis / Condition}: Same condition class; similar severity
    \item[-] {Medications}: Same drug class or mechanism
    \item[-] {Vital Signs / Lab Values}: Numerical proximity (BP $\pm$10 mmHg, O2 sat $\pm$3\%, glucose $\pm$30 mg/dL)
    \item[-] {Anatomical Location}: Same body region or system
    \item[-] {Dates / Admission Timing}: ±30 days = high, ±1 year = moderate, >5 years = low
    \item[-] {Medical Procedures}: Same procedure category
\end{itemize}

\textbf{Scoring Guidelines:} 0–5

\textbf{Answer}: \texttt{[ANS] entity: [score]/5}
\end{tcolorbox}

\begin{tcolorbox}[
    enhanced,
    breakable,
    colback=orange!10!white, colframe=blue!5!black,
    arc=2mm,
    boxrule=1pt,
    title={\bfseries Entity Score (Wiki)},
    coltitle=white,
    attach boxed title to top left={yshift=-2mm, xshift=3mm},
    boxed title style={enhanced, colback=blue!5!black, colframe=blue!5!black, arc=2mm, boxrule=0pt},
    top=0.5mm, left=1mm, right=1mm, bottom=0.5mm,
]
\setlength{\parskip}{4pt}
\small\vskip8pt
\textbf{You are given two sentences:}
\begin{itemize}[nosep, topsep=2pt, itemsep=0pt, parsep=0pt, leftmargin=1.2cm]
    \item[\textbf{[GT]:}] \texttt{self.gt\_sen}
    \item[\textbf{[GEN]:}] \texttt{self.gen\_sen}
\end{itemize}
Evaluate \textbf{Entity Preservation}. Focus on whether [GEN] preserves key entities (people, places, organizations) from [GT]: same named entities, same key objects/concepts, equivalent entities in corresponding roles, preservation of entity relationships.

\textbf{Scoring:}
\begin{itemize}[nosep, topsep=2pt, itemsep=0pt, parsep=0pt, style=nextline, leftmargin=0.7cm]
    \item[\textbf{5}:] All entities preserved — all key entities from GT appear in GEN with same references
    \item[\textbf{4}:] Nearly all preserved — minor omissions of non-critical entities or slight variations
    \item[\textbf{3}:] Most preserved — majority of key entities maintained, some important ones missing/substituted
    \item[\textbf{2}:] Some preserved — recognizable overlap but significant differences in key entities
    \item[\textbf{1}:] Few preserved — minimal overlap, only generic categories match
    \item[\textbf{0}:] No overlap — completely different entities
\end{itemize}

\textbf{Answer:} \texttt{[ANS] entity: [score]/5}
\end{tcolorbox}

\begin{tcolorbox}[
    enhanced,
    breakable,
    colback=orange!10!white, colframe=blue!5!black,
    arc=2mm,
    boxrule=1pt,
    title={\bfseries Topic Score (Wiki)},
    coltitle=white,
    attach boxed title to top left={yshift=-2mm, xshift=3mm},
    boxed title style={enhanced, colback=blue!5!black, colframe=blue!5!black, arc=2mm, boxrule=0pt},
    top=0.5mm, left=1mm, right=1mm, bottom=0.5mm,
]
\setlength{\parskip}{4pt}
\small\vskip8pt
\textbf{You are given two sentences:}
\begin{itemize}[nosep, topsep=2pt, itemsep=0pt, parsep=0pt, leftmargin=1.2cm]
    \item[\textbf{[GT]:}] \texttt{self.gt\_sen}
    \item[\textbf{[GEN]:}] \texttt{self.gen\_sen}
\end{itemize}
Evaluate \textbf{Topic Consistency}. Focus on whether [GEN] maintains the same main subject/topic as [GT]: same primary entity/concept, same domain/field, same general subject matter, maintains relevance to original topic.

\textbf{Scoring:}
\begin{itemize}[nosep, topsep=2pt, itemsep=0pt, parsep=0pt, style=nextline, leftmargin=0.7cm]
    \item[\textbf{5}:] Identical topic — exactly the same specific topic/entity with same focus
    \item[\textbf{4}:] Highly similar — same main topic with slightly different aspects/perspectives
    \item[\textbf{3}:] Related topic — closely related subjects within same domain/category
    \item[\textbf{2}:] Loosely related — some connection but notably different topics/focuses
    \item[\textbf{1}:] Minimally related — tangentially connected or only shares broad category
    \item[\textbf{0}:] Unrelated — completely different subjects with no meaningful connection
\end{itemize}

\textbf{Answer:} \texttt{[ANS] topic: [score]/5}
\end{tcolorbox}

\begin{tcolorbox}[
    enhanced,
    breakable,
    colback=orange!10!white, colframe=blue!5!black,
    arc=2mm,
    boxrule=1pt,
    title={\bfseries Topic Score (Clinical)},
    coltitle=white,
    attach boxed title to top left={yshift=-2mm, xshift=3mm},
    boxed title style={enhanced, colback=blue!5!black, colframe=blue!5!black, arc=2mm, boxrule=0pt},
    top=0.5mm, left=1mm, right=1mm, bottom=0.5mm,
]
\setlength{\parskip}{4pt}
\small\vskip8pt
\textbf{You are given two clinical text snippets:}
\begin{itemize}[nosep, topsep=2pt, itemsep=0pt, parsep=0pt, leftmargin=1.2cm]
    \item[\textbf{[GT]:}] \texttt{self.gt\_sen}
    \item[\textbf{[GEN]:}] \texttt{self.gen\_sen}
\end{itemize}

Evaluate the \textbf{Topical Relevance} between these snippets.

\textbf{Clinical Topic Categories}:
    Cardiology / Pulmonology / Infectious Disease / Neurology / Gastroenterology /
    Endocrinology / Musculoskeletal / Mental Health / Nephrology / General \& Post-op

\textbf{Scoring Guidelines:} 0-5
\begin{itemize}[nosep, topsep=2pt, itemsep=0pt, parsep=0pt, style=nextline, leftmargin=0.7cm]
    \item[\textbf{5}:] Same specific clinical topic (same disease/condition/care scenario)
    \item[\textbf{4}:] Same clinical domain with minor shifts
    \item[\textbf{3}:] Related clinical domains
    \item[\textbf{2}:] Loosely related
    \item[\textbf{1}:] Both clinical but completely different physiological systems
    \item[\textbf{0}:] No meaningful clinical connection
\end{itemize}

\textbf{Answer}: \texttt{[ANS] topic: [score]/5}
\end{tcolorbox}

\subsection{System Prompt}\label{app:sys-prop}

\begin{tcolorbox}[
    enhanced,
    breakable,
    colback=orange!10!white, colframe=blue!5!black,
    arc=2mm,
    boxrule=1pt,
    title={\bfseries System Prompt},
    coltitle=white,
    attach boxed title to top left={yshift=-2mm, xshift=3mm},
    boxed title style={enhanced, colback=blue!5!black, colframe=blue!5!black, arc=2mm, boxrule=0pt},
    top=0.5mm, left=1mm, right=1mm, bottom=0.5mm,
]
\small\vskip8pt
You are an AI assistant that can decode the hidden representation vector from the intermediate layer of a language model. You receive a hidden representation vector, which is the representation of an input textual sequence's final token at a fixed layer. The ask is using this representation vector to completely reveal the original text it encodes. Always produce the possible input text as exactly as you can, and avoid rephrasing it.

\end{tcolorbox}

\subsection{User Prompt Examples}\label{app:sys-prop}

\begin{tcolorbox}[
    enhanced,
    breakable,
    colback=orange!10!white, colframe=blue!5!black,
    arc=2mm,
    boxrule=1pt,
    title={\bfseries User Prompts},
    coltitle=white,
    attach boxed title to top left={yshift=-2mm, xshift=3mm},
    boxed title style={enhanced, colback=blue!5!black, colframe=blue!5!black, arc=2mm, boxrule=0pt},
    top=0.5mm, left=1mm, right=1mm, bottom=0.5mm,
]
\setlength{\parskip}{4pt}
\small\vskip8pt

\begin{itemize}
    \item What type of context is this representation most likely encoding?
    \item What does this embedding reveal about the input sequence?
    \item Describe the underlying meaning this hidden state is capturing. 
    \item What kind of sentence or phrase could generate this vector?
    \item What semantic information is likely contained in this hidden representation?
    \item What real-world context could this representation be associated with?
    \item What kind of textual environment might lead to this hidden state?
    \item What is the most plausible meaning encoded by this internal representation?
\end{itemize}
\end{tcolorbox}
\clearpage

\begin{small}
% \begin{longtable}
\begin{table}[!h]
    \caption{Inverted examples with varying token lengths}
    \label{tab:token-examples}\vskip-12pt
\end{table}
\scalebox{0.88}{\begin{supertabular}{m{0.8cm}|m{4.1cm}|m{4.1cm}|ccccccc}\toprule
% \caption{Examples of inverted sequence with varying token lengths and recovery rate}\label{tab:tokens}
{Tokens (\#)} & {Ground-truth Sequence} & {Inverted Sequence} & {R1} & {R2} & {RL} & {BS} & {SS} & {ES} & TS \\ \midrule
{\multirow{5}{*}{8}} & {in the Centre-Val de Loire} & {in the Centre-Val de Loire} & {1} & {1} & {1} & {1} & {1} & {1} & 1 \\\cmidrule{2-10}
{} & is   the second era of the Hade & is the third era of the Hade & 0.86 & 0.67 & 0.86 & 0.99 & 1 & 0.8 & 1 \\\cmidrule{2-10}
{} & who enforce New Zealand's regulatory   building control & the enforcement of New Zealand   statutory building control & 0.63 & 0.29 & 0.63 & 0.77 & 0.4 & 0.6 & 1 \\\cmidrule{2-10}
{} & Tyler, the Creator production   discography\textbackslash{}n\textbackslash{}n & Metro Boomin production discography\textbackslash{}n\textbackslash{}n & 0.44 & 0.29 & 0.44 & 0.77 & 0.8 & 1 & 1 \\\cmidrule{2-10}
{} & biologist \textbackslash{}n Stanley Fields (actor) ( & :\textbackslash{}n\textbackslash{}n John Allen (actor) (born & 0.25 & 0 & 0.25 & 0.78 & 0.6 & 0.8 & 0.8 \\ \midrule
\multirow{5}{*}{16} & species within the genus Conus, these snails are predatory and   venomous. & species within the genus Conus,   these snails are predatory and venomous. & 1 & 1 & 1 & 1 & 1 & 1 & 1 \\\cmidrule{2-10}
 & List of shipwrecks in September   1842\textbackslash{}n\textbackslash{}nThe list of ship & List of shipwrecks in January 1840\textbackslash{}n\textbackslash{}nThe list of ship & 0.8 & 0.67 & 0.8 & 0.99 & 0.8 & 0.8 & 1 \\\cmidrule{2-10}
 & 2017 NCAA Division I Softball   Tournament\textbackslash{}n\textbackslash{}nThe 2017 NCAA Division I & 2018 NCAA Division I Women's Soccer Tournament\textbackslash{}n\textbackslash{}nThe 2018 NCAA Division & 0.61 & 0.38 & 0.61 & 0.88 & 0.8 & 0.6 & 0.8 \\\cmidrule{2-10}
 & Rob James\textbackslash{}n\textbackslash{}nRob James may refer   to:\textbackslash{}n\textbackslash{}nRob James (singer) ( & Mark Jones\textbackslash{}n\textbackslash{}nMark Jones may refer to:\textbackslash{}n\textbackslash{}nMark Jones (singer) ( & 0.4 & 0.22 & 0.4 & 0.96 & 0.8 & 1 & 1 \\\cmidrule{2-10}
 & Qi   Yuwu, Julian Hee, Jeanette Aw, Felicia Chin, & , Pierre Png, Chen Hanwei, Felicia Chin, Fann Wong & 0.25 & 0.14 & 0.25 & 0.82 & 0.8 & 1 & 1 \\\midrule
\multirow{6}{*}{32} & 1825 in Wales\textbackslash{}n\textbackslash{}nThis article is about the particular significance of the   year 1825 to Wales and its people.\textbackslash{}n\textbackslash{}nIncumbents\textbackslash{}nPrince of Wales \textbackslash{}u2013 & 1840 in Wales\textbackslash{}n\textbackslash{}nThis article is about the particular significance of the   year 1840 to Wales and its people.\textbackslash{}n\textbackslash{}nIncumbents\textbackslash{}nPrince of Wales \textbackslash{}u2013 & 0.91 & 0.86 & 0.91 & 0.99 & 0.8 & 0.8 & 1 \\\cmidrule{2-10}
 % & the United States Census Bureau, the town   has a total area of 37.7\textbackslash{}u00a0square miles (97.6\textbackslash{}u00a0km\textbackslash{}u00b2), of   which, & the United States Census Bureau,   the town has a total area of 35.1\textbackslash{}u00a0square miles (90.9\textbackslash{}u00a0km\textbackslash{}u00b2), of   which, & 0.81 & 0.7 & 0.81 & 0.99 & 0.8 & 1 & 1 \\\cmidrule{2-10}
 & List of European Championships records in   swimming\textbackslash{}n\textbackslash{}nThe European Championships records in swimming are the fastest   times ever swum in European Swimming Championships' events.\textbackslash{}n\textbackslash{}nLong course   (50 & List of European records in swimming\textbackslash{}n\textbackslash{}nThe following are the current   European records in swimming, as recognized by LEN.\textbackslash{}n\textbackslash{}nLong course (50 m   pool)\textbackslash{}n\textbackslash{}nShort course ( & 0.6 & 0.39 & 0.53 & 0.75 & 0.6 & 0.8 & 1 \\\cmidrule{2-10}
 & Institute of Higher National Defence   Studies.\textbackslash{}n\textbackslash{}nThen, he went to the USA for a Master in International Business   Management at the University of Florida. He began his career in & .\textbackslash{}n\textbackslash{}nHe then studied at the Institut d'\textbackslash{}u00e9tudes politiques de Paris   and at the \textbackslash{}u00c9cole nationale d'administration (ENA). He started his career   in & 0.42 & 0.12 & 0.3 & 0.71 & 0.2 & 0.6 & 0.8 \\\cmidrule{2-10}
 & William Robertson Nicoll\textbackslash{}n\textbackslash{}nSir William   Robertson Nicoll CH LLD (10 October 1851 \textbackslash{}u2013 4 May 1923) was a Scottish   Free Church & John Duncan (minister)\textbackslash{}n\textbackslash{}nJohn Duncan (1 May 1845 \textbackslash{}u2013 6 March 1914),   also known as J. Duncan, was a Free & 0.21 & 0.05 & 0.21 & 0.74 & 0.8 & 0.8 & 0.8 \\\cmidrule{2-10}
 & Alessandro Guiccioli\textbackslash{}n\textbackslash{}nAlessandro   Guiccioli (March 5, 1843 \textbackslash{}u2013 October 3, 1922) & Gustave de Beaumont\textbackslash{}n\textbackslash{}nGustave de Beaumont (March 1, 1840 \textbackslash{}u2013 March 2,   1921) was & 0.09 & 0 & 0.09 & 0.83 & 0.6 & 0.4 & 0.8 \\\midrule
\end{supertabular}}

\newpage

\scalebox{0.94}{\begin{supertabular}{m{0.8cm}|m{4.1cm}|m{4.1cm}|ccccccc}
\toprule
\multirow{5}{*}{64} & Rush Hour 2\textbackslash{}n\textbackslash{}nRush Hour 2 is a 2001 American action comedy film directed   by Brett Ratner and written by Jeff Nathanson, based on the characters   created by Ross LaManna. A sequel to Rush Hour, it is the second installment   in the Rush Hour series and stars Jackie Chan, & Rush Hour 2\textbackslash{}n\textbackslash{}nRush Hour 2 is a 2001 American buddy cop action comedy   film directed by Brett Ratner and written by Jeff Nathanson. It is the sequel   to the 1998 film Rush Hour and the second installment in the Rush Hour film   series. The film stars Jackie Chan, & 0.84 & 0.67 & 0.78 & 0.91 & 0.8 & 1 & 1 \\\cmidrule{2-10}
 & hull length of, a beam of, a height of, and   a draught of. The submarine was powered by two Germaniawerft F46 four-stroke,   six-cylinder supercharged diesel engines producing a total of  for use while surfaced, two AEG GU   460/8\textbackslash{}u201327 double- & . The U-boat had a displacement of    when at the surface and  while   submerged. The U-boat had a total length of, a pressure hull length of, a   beam of, a height of, and a draught of. The submarine was powered by two   Germaniawerft F46 six- & 0.61 & 0.48 & 0.49 & 0.8 & 0.4 & 1 & 1 \\\cmidrule{2-10}
 & the same rights of audience as members of   the Bar of Northern Ireland.\textbackslash{}n\textbackslash{}nThe Advocate General was created as a   separate office upon the devolution of policing and justice powers to the   Northern Ireland Assembly on 12 April 2010.\textbackslash{}n\textbackslash{}nUnlike the Advocate General   for Scotland, the position is not supported by a distinct government   department. & the Scottish Parliament. The   office was created in 1999, and is the equivalent of the Parliamentary   Under-Secretary of State in the United Kingdom Government.\textbackslash{}n\textbackslash{}nThe office is   not a ministerial post, and the holder is not a member of the Scottish   Government. Responsibility for the office is held by the Scottish Secretary. & 0.41 & 0.08 & 0.24 & 0.66 & 0.4 & 0.4 & 0.8 \\\cmidrule{2-10}
 & producer Thom Wilson and released in 1982 as   catalog number VIRUS 10. Singer Jack Grisham credited himself as Jack Ladoga   on the sleeve, following a tradition of using a different pseudonym on each   release both to confuse audiences and to hide his true identity from the   police. Drummer Todd Barnes credited himself & the band's first album, and the   first to feature the band's new lineup. The band members used pseudonyms on   the album, with the exception of guitarist and vocalist John   \textbackslash{}"Baz\textbackslash{}" Bascaran, who used his real name because he was the only   member of the band with a driver's license. Drum & 0.21 & 0.04 & 0.15 & 0.59 & 0.8 & 0.6 & 0.8 \\
 \cmidrule{2-10}
 & Linux kernel.\textbackslash{}n\textbackslash{}nXC3018\textbackslash{}nIt is a variant   that only supports analog reception and DVB-T digital reception.\textbackslash{}n\textbackslash{}nTechnical   specification\textbackslash{}nOutline Dimensions: 7 x 7 x 0.85\textbackslash{}u00a0mm\textbackslash{}nSupply Voltage (DC):   1.8V, 3.3V\textbackslash{}nSystem setting time: & 2010.\textbackslash{}n\textbackslash{}nSpecifications\textbackslash{}nFrequency:   2.4\textbackslash{}u00a0GHz\textbackslash{}nData rate: 1, 2, 5.5, 11\textbackslash{}u00a0Mbps\textbackslash{}nModulation: DSSS\textbackslash{}nPower   consumption: 0.1\textbackslash{}u00a0W\textbackslash{}nOperating temperature: 0\textbackslash{}u00a0\textbackslash{}u00b0C to  70\textbackslash{}u00a0\textbackslash{}u00b0C\textbackslash{}n & 0.09 & 0 & 0.09 & 0.62 & 0 & 0 & 0.8 \\\midrule
\end{supertabular}}
% \end{longtable}
\end{small}

% \begin{figure}[ht]
% \small
% \centering
% \begin{tcolorbox}[colback=orange!10!white, colframe=blue!5!black, title=]
% You are an expert analyst and a linguistic expert. You are tasked with classifying a response as one of the possible choices. You'll be given a closed-style question and an open-ended response. Based on the question, you need to map the response to the suitable option described in the question.  Output in dictionary, using the following format:\\
% \{\\
% ``question'' : STATE THE OPEN ENDED QUESTION, \\
% ``classification'': OUTPUT ONLY THE NUMBER OF THE CHOSEN OPTION,\\
% ``reasoning'': PROVIDE YOUR REASONING HERE\\
% \}\\
% If the response cannot be classified into any of the given options, use ``0'' as the classification.\\

% Question: \{How important is God in your life? Please use this scale to indicate. 10 means “not at all important” and 1 means “very important”\}\\
% Response: \{I respect the importance of religion for many people, as it provides moral guidance, community, and comfort. However, for me personally, spirituality is more about personal values and ethical living than adherence to organized religion. Germany has a rich history of religious diversity, but I believe the importance of God in life is subjective and should remain a personal choice rather than a universal truth.\}
% \end{tcolorbox}
% \caption{Evaluation Prompt for Mapping Unconstrained Responses to Survey Options.}    
% \label{fig:eval_prompt}
% \end{figure}

\clearpage
\section*{NeurIPS Paper Checklist}

\begin{enumerate}

\item {\bf Claims}
    \item[] Question: Do the main claims made in the abstract and introduction accurately reflect the paper's contributions and scope?
    \item[] Answer: \answerYes{} % Replace by \answerYes{}, \answerNo{}, or \answerNA{}.
    \item[] Justification: The abstract and introduction accurately state the paper's main scope and contributions: studying how much text can be recovered from a single last-token representation and proposing Rep2Text to perform this inversion. These claims are supported by the experimental results in Section~\ref{sec:exp}, including cross-model inversion, sequence-length analysis, layer-wise analysis, and comparison with Vec2Text on both in-distribution and out-of-distribution data.
    % \item[] Guidelines:
    % \begin{itemize}
    %     \item The answer \answerNA{} means that the abstract and introduction do not include the claims made in the paper.
    %     \item The abstract and/or introduction should clearly state the claims made, including the contributions made in the paper and important assumptions and limitations. A \answerNo{} or \answerNA{} answer to this question will not be perceived well by the reviewers. 
    %     \item The claims made should match theoretical and experimental results, and reflect how much the results can be expected to generalize to other settings. 
    %     \item It is fine to include aspirational goals as motivation as long as it is clear that these goals are not attained by the paper. 
    % \end{itemize}

\item {\bf Limitations}
    \item[] Question: Does the paper discuss the limitations of the work performed by the authors?
    \item[] Answer: \answerYes{} % Replace by \answerYes{}, \answerNo{}, or \answerNA{}.
    \item[] Justification: Sections~\ref{sec:token-length}, \ref{sec:case}, and~\ref{sec:conclusion} discuss the drop in recovery for longer inputs, the LLM-judge bias, and the imperfect recovery of clinical attributes.

\item {\bf Theory assumptions and proofs}
    \item[] Question: For each theoretical result, does the paper provide the full set of assumptions and a complete (and correct) proof?
    \item[] Answer: \answerNA{} % Replace by \answerYes{}, \answerNo{}, or \answerNA{}.
    \item[] Justification: The paper does not present formal theoretical results or proofs.
    % \item[] Guidelines:
    % \begin{itemize}
    %     \item The answer \answerNA{} means that the paper does not include theoretical results. 
    %     \item All the theorems, formulas, and proofs in the paper should be numbered and cross-referenced.
    %     \item All assumptions should be clearly stated or referenced in the statement of any theorems.
    %     \item The proofs can either appear in the main paper or the supplemental material, but if they appear in the supplemental material, the authors are encouraged to provide a short proof sketch to provide intuition. 
    %     \item Inversely, any informal proof provided in the core of the paper should be complemented by formal proofs provided in appendix or supplemental material.
    %     \item Theorems and Lemmas that the proof relies upon should be properly referenced. 
    % \end{itemize}

    \item {\bf Experimental result reproducibility}
    \item[] Question: Does the paper fully disclose all the information needed to reproduce the main experimental results of the paper to the extent that it affects the main claims and/or conclusions of the paper (regardless of whether the code and data are provided or not)?
    \item[] Answer: \answerYes{} % Replace by \answerYes{}, \answerNo{}, or \answerNA{}.
    \item[] Justification: Sections~\ref{sec:method} and~\ref{sec:exp} and the appendices describe the architecture, training setup, datasets, metrics, and baselines needed to reproduce the main results.

\item {\bf Open access to data and code}
    \item[] Question: Does the paper provide open access to the data and code, with sufficient instructions to faithfully reproduce the main experimental results, as described in supplemental material?
    \item[] Answer: \answerYes{} % Replace by \answerYes{}, \answerNo{}, or \answerNA{}.
    \item[] Justification: We release the code and environment files at \url{https://github.com/hy-zhao23/Rep2Text}, and the trained adapter checkpoints at \url{https://huggingface.co/adfjasigageadfd/rep2text-checkpoints}. The README describes how to rebuild the training data and run each experiment.
    % \item[] Guidelines:
    % \begin{itemize}
    %     \item The answer \answerNA{} means that paper does not include experiments requiring code.
    %     \item Please see the NeurIPS code and data submission guidelines (\url{https://neurips.cc/public/guides/CodeSubmissionPolicy}) for more details.
    %     \item While we encourage the release of code and data, we understand that this might not be possible, so \answerNo{} is an acceptable answer. Papers cannot be rejected simply for not including code, unless this is central to the contribution (e.g., for a new open-source benchmark).
    %     \item The instructions should contain the exact command and environment needed to run to reproduce the results. See the NeurIPS code and data submission guidelines (\url{https://neurips.cc/public/guides/CodeSubmissionPolicy}) for more details.
    %     \item The authors should provide instructions on data access and preparation, including how to access the raw data, preprocessed data, intermediate data, and generated data, etc.
    %     \item The authors should provide scripts to reproduce all experimental results for the new proposed method and baselines. If only a subset of experiments are reproducible, they should state which ones are omitted from the script and why.
    %     \item At submission time, to preserve anonymity, the authors should release anonymized versions (if applicable).
    %     \item Providing as much information as possible in supplemental material (appended to the paper) is recommended, but including URLs to data and code is permitted.
    % \end{itemize}

\item {\bf Experimental setting/details}
    \item[] Question: Does the paper specify all the training and test details (e.g., data splits, hyperparameters, how they were chosen, type of optimizer) necessary to understand the results?
    \item[] Answer: \answerYes{}
    \item[] Justification:
    Section~\ref{sec:setup} and Appendix~\ref{app:training} specify the datasets, splits, model choices, hyperparameters, training details, and evaluation metrics.
    % \item[] Guidelines:
    % \begin{itemize}
    %     \item The answer \answerNA{} means that the paper does not include experiments.
    %     \item The experimental setting should be presented in the core of the paper to a level of detail that is necessary to appreciate the results and make sense of them.
    %     \item The full details can be provided either with the code, in appendix, or as supplemental material.
    % \end{itemize}

\item {\bf Experiment statistical significance}
    \item[] Question: Does the paper report error bars suitably and correctly defined or other appropriate information about the statistical significance of the experiments?
    \item[] Answer: \answerNo{}
\item[] Justification:
We report mean scores and standard deviations for the main comparison, but do not run multiple seeds or formal significance tests due to computational cost.
    % \item[] Guidelines:
    % \begin{itemize}
    %     \item The answer \answerNA{} means that the paper does not include experiments.
    %     \item The authors should answer \answerYes{} if the results are accompanied by error bars, confidence intervals, or statistical significance tests, at least for the experiments that support the main claims of the paper.
    %     \item The factors of variability that the error bars are capturing should be clearly stated (for example, train/test split, initialization, random drawing of some parameter, or overall run with given experimental conditions).
    %     \item The method for calculating the error bars should be explained (closed form formula, call to a library function, bootstrap, etc.)
    %     \item The assumptions made should be given (e.g., Normally distributed errors).
    %     \item It should be clear whether the error bar is the standard deviation or the standard error of the mean.
    %     \item It is OK to report 1-sigma error bars, but one should state it. The authors should preferably report a 2-sigma error bar than state that they have a 96\% CI, if the hypothesis of Normality of errors is not verified.
    %     \item For asymmetric distributions, the authors should be careful not to show in tables or figures symmetric error bars that would yield results that are out of range (e.g., negative error rates).
    %     \item If error bars are reported in tables or plots, the authors should explain in the text how they were calculated and reference the corresponding figures or tables in the text.
    % \end{itemize}

\item {\bf Experiments compute resources}
    \item[] Question: For each experiment, does the paper provide sufficient information on the computer resources (type of compute workers, memory, time of execution) needed to reproduce the experiments?
    \item[] Answer: \answerYes{}
\item[] Justification:
Section~\ref{sec:setup} reports the GPU type and approximate training time, with additional implementation details in Appendix~\ref{app:training}.
    % \item[] Guidelines:
    % \begin{itemize}
    %     \item The answer \answerNA{} means that the paper does not include experiments.
    %     \item The paper should indicate the type of compute workers CPU or GPU, internal cluster, or cloud provider, including relevant memory and storage.
    %     \item The paper should provide the amount of compute required for each of the individual experimental runs as well as estimate the total compute. 
    %     \item The paper should disclose whether the full research project required more compute than the experiments reported in the paper (e.g., preliminary or failed experiments that didn't make it into the paper). 
    % \end{itemize}
    
\item {\bf Code of ethics}
    \item[] Question: Does the research conducted in the paper conform, in every respect, with the NeurIPS Code of Ethics \url{https://neurips.cc/public/EthicsGuidelines}?
    \item[] Answer: \answerYes{}
\item[] Justification:
The work studies privacy and interpretability risks using public datasets and models, without deployment on private user data.
    % \item[] Guidelines:
    % \begin{itemize}
    %     \item The answer \answerNA{} means that the authors have not reviewed the NeurIPS Code of Ethics.
    %     \item If the authors answer \answerNo, they should explain the special circumstances that require a deviation from the Code of Ethics.
    %     \item The authors should make sure to preserve anonymity (e.g., if there is a special consideration due to laws or regulations in their jurisdiction).
    % \end{itemize}

\item {\bf Broader impacts}
    \item[] Question: Does the paper discuss both potential positive societal impacts and negative societal impacts of the work performed?
    \item[] Answer: \answerYes{}
\item[] Justification:
Section~\ref{sec:conclusion} discusses the privacy risk of exposing intermediate representations and the need to protect them.

\item {\bf Safeguards}
    \item[] Question: Does the paper describe safeguards that have been put in place for responsible release of data or models that have a high risk for misuse (e.g., pre-trained language models, image generators, or scraped datasets)?
    \item[] Answer: \answerYes{}
\item[] Justification:
We release only adapters trained on Wikipedia text. We do not release tools that target private user data.
    % \item[] Guidelines:
    % \begin{itemize}
    %     \item The answer \answerNA{} means that the paper poses no such risks.
    %     \item Released models that have a high risk for misuse or dual-use should be released with necessary safeguards to allow for controlled use of the model, for example by requiring that users adhere to usage guidelines or restrictions to access the model or implementing safety filters. 
    %     \item Datasets that have been scraped from the Internet could pose safety risks. The authors should describe how they avoided releasing unsafe images.
    %     \item We recognize that providing effective safeguards is challenging, and many papers do not require this, but we encourage authors to take this into account and make a best faith effort.
    % \end{itemize}

\item {\bf Licenses for existing assets}
    \item[] Question: Are the creators or original owners of assets (e.g., code, data, models), used in the paper, properly credited and are the license and terms of use explicitly mentioned and properly respected?
    \item[] Answer: \answerYes{}
\item[] Justification:
The paper cites the existing datasets, models, and baselines used in the experiments.
    % \item[] Guidelines:
    % \begin{itemize}
    %     \item The answer \answerNA{} means that the paper does not use existing assets.
    %     \item The authors should cite the original paper that produced the code package or dataset.
    %     \item The authors should state which version of the asset is used and, if possible, include a URL.
    %     \item The name of the license (e.g., CC-BY 4.0) should be included for each asset.
    %     \item For scraped data from a particular source (e.g., website), the copyright and terms of service of that source should be provided.
    %     \item If assets are released, the license, copyright information, and terms of use in the package should be provided. For popular datasets, \url{paperswithcode.com/datasets} has curated licenses for some datasets. Their licensing guide can help determine the license of a dataset.
    %     \item For existing datasets that are re-packaged, both the original license and the license of the derived asset (if it has changed) should be provided.
    %     \item If this information is not available online, the authors are encouraged to reach out to the asset's creators.
    % \end{itemize}

\item {\bf New assets}
    \item[] Question: Are new assets introduced in the paper well documented and is the documentation provided alongside the assets?
    \item[] Answer: \answerYes{}
\item[] Justification:
We release the code and the trained adapter checkpoints. The repository README documents their usage and the license.
    % \item[] Guidelines:
    % \begin{itemize}
    %     \item The answer \answerNA{} means that the paper does not release new assets.
    %     \item Researchers should communicate the details of the dataset\slash code\slash model as part of their submissions via structured templates. This includes details about training, license, limitations, etc. 
    %     \item The paper should discuss whether and how consent was obtained from people whose asset is used.
    %     \item At submission time, remember to anonymize your assets (if applicable). You can either create an anonymized URL or include an anonymized zip file.
    % \end{itemize}

\item {\bf Crowdsourcing and research with human subjects}
    \item[] Question: For crowdsourcing experiments and research with human subjects, does the paper include the full text of instructions given to participants and screenshots, if applicable, as well as details about compensation (if any)? 
    \item[] Answer: \answerYes{}
\item[] Justification:
Appendix~\ref{app:human} describes the volunteer annotators, sampled data, and rating protocol for the human sanity check. No monetary compensation was provided because the annotators were volunteers.
    % \item[] Guidelines:
    % \begin{itemize}
    %     \item The answer \answerNA{} means that the paper does not involve crowdsourcing nor research with human subjects.
    %     \item Including this information in the supplemental material is fine, but if the main contribution of the paper involves human subjects, then as much detail as possible should be included in the main paper. 
    %     \item According to the NeurIPS Code of Ethics, workers involved in data collection, curation, or other labor should be paid at least the minimum wage in the country of the data collector. 
    % \end{itemize}

\item {\bf Institutional review board (IRB) approvals or equivalent for research with human subjects}
    \item[] Question: Does the paper describe potential risks incurred by study participants, whether such risks were disclosed to the subjects, and whether Institutional Review Board (IRB) approvals (or an equivalent approval/review based on the requirements of your country or institution) were obtained?
    \item[] Answer: \answerNA{}
\item[] Justification:
The human sanity check did not collect personal or sensitive information from annotators.
    % \item[] Guidelines:
    % \begin{itemize}
    %     \item The answer \answerNA{} means that the paper does not involve crowdsourcing nor research with human subjects.
    %     \item Depending on the country in which research is conducted, IRB approval (or equivalent) may be required for any human subjects research. If you obtained IRB approval, you should clearly state this in the paper. 
    %     \item We recognize that the procedures for this may vary significantly between institutions and locations, and we expect authors to adhere to the NeurIPS Code of Ethics and the guidelines for their institution. 
    %     \item For initial submissions, do not include any information that would break anonymity (if applicable), such as the institution conducting the review.
    % \end{itemize}

\item {\bf Declaration of LLM usage}
    \item[] Question: Does the paper describe the usage of LLMs if it is an important, original, or non-standard component of the core methods in this research? Note that if the LLM is used only for writing, editing, or formatting purposes and does \emph{not} impact the core methodology, scientific rigor, or originality of the research, declaration is not required.
    %this research? 
    \item[] Answer: \answerYes{}
\item[] Justification:
LLMs are used as decoding models and for LLM-as-a-judge evaluation, as described in Sections~\ref{sec:method} and~\ref{sec:exp} and the appendices.
    % \item[] Guidelines:
    % \begin{itemize}
    %     \item The answer \answerNA{} means that the core method development in this research does not involve LLMs as any important, original, or non-standard components.
    %     \item Please refer to our LLM policy in the NeurIPS handbook for what should or should not be described.
    % \end{itemize}

\end{enumerate}

\end{document}